\documentclass[useAMS,usegraphicx,referee]{biom}
\def\bSig\mathbf{\Sigma}

\usepackage[utf8]{inputenc}
\usepackage{url}
\usepackage{xurl}

\usepackage{caption}
\usepackage{subcaption}

\usepackage{graphicx}
\usepackage[english]{babel}
\usepackage[nopatch=footnote]{microtype}
\usepackage[dvipsnames]{xcolor}
\usepackage{algorithm}
\usepackage{algorithmicx}
\usepackage{amsmath}
\usepackage{amssymb}
\usepackage{algpseudocode}
\usepackage{hyperref}
\usepackage{setspace}
\usepackage{multirow}

\let\bibhang\relax
\usepackage[backend=biber,sorting=none,style=apa]{biblatex}
\title[Robust Variable Selection via Group Testing]{Robust Ultra-High-Dimensional Variable Selection With Correlated Structure Using Group Testing}

\author
{Wanru Guo\emailx{wanru.guo@som.umaryland.edu} \\
Department of Biostatistics and Bioinformatics
University of Maryland, Baltimore \\ Marlene \& Stewart Greenebaum Comprehensive Cancer Center,
Baltimore, Maryland, U.S.A.
\and
Juan Xie\emailx{juan.xie@som.umaryland.edu} \\
Department of Biostatistics and Bioinformatics
University of Maryland, Baltimore \\ Marlene \& Stewart Greenebaum Comprehensive Cancer Center,
Baltimore, Maryland, U.S.A.
\and
Binbin Wang\emailx{binbin.wang@som.umaryland.edu} \\
Department of Biostatistics and Bioinformatics
University of Maryland, Baltimore \\ Marlene \& Stewart Greenebaum Comprehensive Cancer Center,
Baltimore, Maryland, U.S.A.
\and
Weicong Chen\emailx{weicong@case.edu} \\
Department of Computer and Data Sciences \\
Case Western Reserve University,
Cleveland, Ohio, U.S.A.
\and
Xiaoyi Lu\emailx{xiaoyilu@ufl.edu} \\
Department of Electrical \& Computer Engineering \\
University of Florida,
Gainesville, Florida, U.S.A.
\and
Vipin Chaudhary\emailx{vipin@case.edu} \\
Department of Computer and Data Sciences \\
Case Western Reserve University,
Cleveland, Ohio, U.S.A.
\and
Curtis Tatsuoka$^\ast$\emailx{ctatsuoka@som.umaryland.edu} \\
Department of Biostatistics and Bioinformatics
University of Maryland, Baltimore \\ Marlene \& Stewart Greenebaum Comprehensive Cancer Center,
Baltimore, Maryland, U.S.A.
}

\begin{document}
\raggedbottom
\label{firstpage}

\begin{abstract}
\\\textbf{Background:} High-dimensional genomic data exhibit inherent group correlation structures that challenge conventional feature selection methods. Traditional approaches treat features as independent, while existing group-regularized methods require pre-specified pathways and suffer from high false discovery rates, computational complexity, and outlier sensitivity.

\noindent\textbf{Methods:} We introduce the Dorfman screening framework, a multi-stage hierarchical procedure that: (1) constructs data-driven groups via hierarchical clustering using sparse correlation (graphical LASSO) or robust measures (OGK covariance, Spearman); (2) performs two-stage hypothesis testing: global group-level tests followed by within-group variable screening; (3) refines selection via elastic net or adaptive elastic net; and (4) incorporates robustness through Huber-weighted regression and dynamic tree cutting for contaminated data.

\noindent\textbf{Results:} Simulations (n=200, p=1000, 200 groups) with group structure of varying correlations and coefficients show that Dorfman-Sparse-Adaptive-EN excels in normal data (F1 = 0.926, RMSE = 1.256), with Robust-OGK-Dorfman-Adaptive-EN as the runner-up (F1 = 0.925, RMSE = 1.641). Under the same simulation but corrupted with broken correlations, batch effects, and asymmetric noise, Robust-OGK-Dorfman-Adaptive-EN (F1 = 0.809, RMSE = 5.249) demonstrates a clear advantage, significantly outperforming classical Dorfman methods and all competitors, including adaptive EN, EN, SIS-LASSO, and gAR2. Applied to GDSC RNA-seq data for trametinib response in NSCLC, robust Dorfman methods achieved superior predictive accuracy (RMSE = 2.17–2.33) and enriched selection of clinically relevant genes.

\noindent\textbf{Conclusions:} The Dorfman framework provides a robust, efficient solution for genomic biomarker discovery. Robust-OGK-Dorfman-Adaptive-EN emerges as the method of choice, performing competitively in normal scenarios while offering decisive improvements under realistic data contamination and non-normality. The framework is scalable to ultra-high dimensions (p=100,000) via block estimation and Bayesian optimization, ensuring applicability to next-generation genomic studies.

\end{abstract}

\begin{keywords}
High-dimensional statistics; group testing; robust covariance estimation; graphical LASSO; elastic net; structured feature selection; genomics; drug sensitivity prediction.
\end{keywords}

\maketitle

\section{Introduction}

High-dimensional biomedical data generated by modern platforms such as multi-omics profiling, high-resolution imaging, and electronic health records continue to expand rapidly in both scale and complexity. A central challenge in these settings is variable screening for regression modeling, wherein the number of candidate predictors ($p$) far exceeds the sample size ($n$) (\cite{fan2008sure,buhlmann2011statistics}). In practice, only a small fraction of variables is truly associated with the outcome of interest, while the vast majority are irrelevant, resulting in a classic ``needle-in-a-haystack’’ problem (\cite{fan2009ultrahigh}).

To address this challenge, a broad class of high-dimensional screening and selection methods has been developed, including sure independence screening (SIS), the LASSO, and the elastic net (\cite{tibshirani1996regression,zou2005regularization,fan2008sure}). SIS established the sure screening property, showing that marginal correlation–based screening can retain all relevant predictors with high probability even in ultra-high-dimensional regimes where $\log(p)=O(n^{\alpha})$ for $\alpha\in(0,1/2)$ (\cite{fan2008sure}). These ideas motivated extensive development of both model-based and model-free screening procedures (\cite{fan2009ultrahigh,fan2010sure,wang2009forward,li2012feature,he2013quantile,Obozinski_2011,zhao2012principled}). However, these approaches fundamentally operate on individual predictors and implicitly treat features as marginally independent.

Such assumptions are routinely violated in genomic and biomedical data, where genes and molecular features exhibit strong group-wise correlation structures driven by shared biological pathways, regulatory networks, and functional modules (\cite{barabasi2004network,subramanian2005gene}). Ignoring these dependencies can lead to unstable feature selection and failure to recover biologically coherent signals in ultra-high-dimensional settings (\cite{zou2005regularization,meinshausen2010stability}). Group-regularized extensions, including the group LASSO, sparse group LASSO, and group-based screening methods such as group SIS, AR2, and HOLP, have been proposed to address correlated predictors (\cite{yuan2006model,huang2012selective,simon2013sparse,qiu2020grouped}). However, these methods rely on prespecified group or pathway annotations, which are often incomplete, context-dependent, or unavailable, and can result in inflated false discovery rates when group definitions are misspecified (\cite{jacob2009group,buhlmann2014high}). Moreover, group-based regularization can be computationally demanding in ultra-high dimensions and remains sensitive to outliers and distributional departures, limiting robustness in large-scale omics applications (\cite{fan2020statistical,alfons2013sparse}).

Group testing provides an alternative paradigm for efficient screening under sparsity. Originally proposed by Dorfman in the context of infectious disease testing (\cite{dorfman1943detection}), group testing exploits low prevalence by pooling samples, allowing many negatives to be cleared with a single test and substantially reducing cost. When a pooled test is positive, follow-up individual testing is performed, yielding large efficiency gains in sparse settings. Modern extensions have incorporated probabilistic and Bayesian formulations that address practical challenges such as dilution effects and measurement error, further improving inferential performance in biomedical applications (\cite{tatsuoka2023bayesian}).

Motivated by these ideas, we introduce the Dorfman variable screening framework for continuous outcomes, a computationally efficient multi-stage hierarchical procedure that adapts group-testing principles to high-dimensional regression. The procedure first clusters genes into data-driven, biologically meaningful groups using hierarchical clustering based on Pearson correlation or sparse correlation estimated via the graphical LASSO, which induces sparsity in the inverse covariance matrix and stabilizes clustering in high dimensions (\cite{friedman2008sparse}). Group-level global tests are then performed to identify candidate signal groups, followed by screenings of individual variables within the identified groups. A final regularized selection step using the elastic net or LASSO refines variable selection while accounting for residual correlation structure (\cite{tibshirani1996regression,zou2005regularization}). Robust variants of the Dorfman framework further address outlier contamination by incorporating Huber-weighted regression throughout group testing and within-group screening (\cite{huber1992robust}), with groups determined by dynamic cut hierarchical clustering (\cite{LangfelderZhangHorvath2008DynamicTreeCut}) using Spearman correlation or sparse correlation derived from an OGK (orthogonalized Gnanadesikan–Kettenring) covariance estimator (\cite{ollerer2015robust}). Together, within this modular structure, these methods provide a unified, scalable, and robust approach to screening grouped predictors in ultra-high-dimensional biomedical data, closely aligned with modern developments in group testing and Bayesian extensions that account for noise and dilution effects (\cite{tatsuoka2023bayesian}). The focus here will be on continuous outcomes in linear regression models, including cases where deviations from classical assumptions arise. Simulations demonstrate the advantages of our proposed Dorfman-based framework, particularly in non-normal data settings. Finally, a real-data example illustrates how the methods are used to identify genes that predict IC50 for an MEK inhibitor in non-small cell lung cancer using bulk RNA sequencing data.
\section{Grouped variable screening}

A key novel feature of our proposed framework for variable screening in ultra-high dimensions is the use of group testing principles to screen groups of variables. In general, when pooled testing is feasible, group testing is most effective when the target binary characteristic has low prevalence (\cite{dorfman1943detection}). The objective of variable screening is to identify all variables that have a significant association with the outcome variable in a regression model (at a specified $\alpha$ level). In variable screening using group testing, pooled tests are regression models comprised of subsets (``pools'') of variables. In our setting, pooled tests are considered ``negative'' when global tests comparing null models (intercept-only) with models that include all variables selected for pooling are not statistically significant. If the global test rejects the null model, the significance of individual variables in the model fit is assessed at the specified $\alpha$ level.

To encourage correct ``positive'' decisions, we first apply hierarchical clustering to form pools of correlated variables, assuming that positive variables are likely to be correlated (e.g., genes are likely to belong to the same pathway), thereby increasing the power to reject the null model (\cite{barabasi2004network,subramanian2005gene}). Conversely, assuming that negative variables also cluster, selecting pools in this manner reduces the Type I error rate for the global test of the pool. This is the critical efficiency advantage of group testing: to efficiently identify negative variables through pooling (\cite{dorfman1943detection}). Selecting individual variables from the pooled tests is inherently vulnerable to Type I errors due to multiple comparisons. Hence, a final stage of this process is to apply penalized regression methods to the set of individual variables selected from the positive pooled tests. Note that the penalized regression at this stage of the process is applied on a smaller set of variables than directly applying penalized regression on the complete set of candidate variables (\cite{tibshirani1996regression,zou2005regularization}).

In sum, there are four major stages to this approach: (1) computing correlations among variables; (2) hierarchical clustering; (3) pooled testing of clusters through regression with the outcome variable; and (4) penalized regression on variables identified as potentially positive from stage 3. The overall approach is outlined in Figure~1.

\begin{figure}[htbp]
    \centering
    \includegraphics[width=0.7\linewidth]{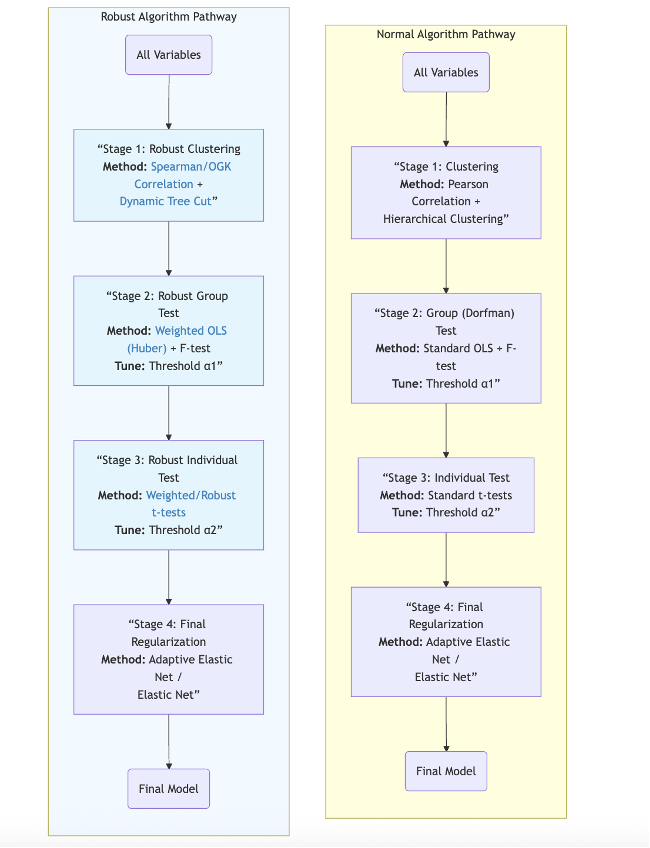}
    \caption{Streamlined Flow Diagram of Normal and Robust Dorfman Procedures.}
    \label{fig:1}
\end{figure}

\subsection{Dorfman–Elastic Net/Adaptive EN with Pearson/Sparse Co-variance and Cross-Validated Cut Height}

Note that the methodologies used at each stage can be replaced without affecting the others. The framework is therefore modular, allowing it to be readily adapted to robust settings, for instance, when the outcome variable is not normally distributed, when variables are skewed or exhibit high kurtosis, or when outliers or influential observations are present. To achieve robustness, we incorporate: (1) robust and sparse correlation estimation (\cite{maronna2002robust,friedman2008sparse}); (2) dynamic tree-cut methods for hierarchical clustering (\cite{LangfelderZhangHorvath2008DynamicTreeCut}); (3) Huber-based robust regression (\cite{huber1992robust}); and (4) elastic net and adaptive elastic net regularization (\cite{zou2005regularization,zou2006adaptive}). Together, these components yield multiple practical variants of group-testing-based approaches for variable selection.

Let $(\mathbf{X}, \mathbf{y})$ denote the $n \times p$ design matrix and response vector, with columns of $\mathbf{X}$ standardized. The Dorfman--EN procedure consists of three stages: (i) construction of a sparse covariance structure and correlation-based grouping via hierarchical clustering with a cross-validated cut height, (ii) Dorfman-style group and variable screening, and (iii) a final penalized regression fit (elastic net or adaptive elastic net).

\noindent\textbf{Stage 1: Pearson/Sparse covariance and correlation-based grouping.}
\textit{Option A (Pearson correlation).}
We compute the empirical Pearson correlation matrix $\hat{R}$ of $\mathbf{X}$ and define the dissimilarity
\[
D_{jk} = 1 - |\hat{R}_{jk}|.
\]
Hierarchical clustering with average linkage is applied to $D$, and the resulting dendrogram is cut at a prespecified height $h$ to obtain covariate groups. 

\textit{Option B (Sparse correlation refinement).}
To stabilize grouping in high-dimensional settings, we estimate a sparse correlation structure using the graphical lasso:
\begin{enumerate}
    \item Compute the empirical covariance matrix $\hat{\Sigma}$ of $\mathbf{X}$.
    \item For a grid of penalty parameters $\rho \in \mathcal{R}$, fit the graphical lasso estimator
    \[
        \hat{\Theta}(\rho)
        = \arg\min_{\Theta \succ 0}
        \big\{ \mathrm{tr}(\hat{\Sigma}\Theta) - \log\det(\Theta) + \rho \|\Theta\|_1 \big\},
    \]
    yielding a sparse precision matrix.
    \item Invert and rescale $\hat{\Theta}(\rho)$ to obtain a sparse correlation matrix
    \[
        \hat{R}^{\mathrm{sp}}(\rho)
        = \mathrm{diag}(\hat{\Theta}(\rho)^{-1})^{-1/2}
          \hat{\Theta}(\rho)^{-1}
          \mathrm{diag}(\hat{\Theta}(\rho)^{-1})^{-1/2}.
    \]
    \item Construct the dissimilarity matrix
    \[
        D_{jk}(\rho) = 1 - \big| \hat{R}^{\mathrm{sp}}_{jk}(\rho) \big|
    \]
    and perform hierarchical clustering with average linkage to obtain a dendrogram.
    \item The penalty parameter $\rho^\star$ is selected by maximizing clustering purity (ARI) over $\mathcal{R}$.
    For the selected $\rho^\star$, the dendrogram is cut at height $h$, where $h$ is chosen by $5$-fold cross-validation
    to minimize prediction RMSE from subsequent screening and EN fitting.
\end{enumerate}
The resulting clusters define candidate groups of covariates. 

\noindent\textbf{Stage 2: Dorfman-style group and individual screening.}
Given the fixed groups $\{g_j\}$ determined at $h^\star$, we apply a two-level screening procedure:
\begin{enumerate}
    \item For each group $g$, let $S_g = \{ j : g_j = g \}$ and denote by $\mathbf{X}_{S_g}$ the corresponding submatrix. We fit the linear model
    \[
        y_i = \beta_{0g} + \mathbf{x}_{i,S_g}^\top \boldsymbol{\beta}_g + \varepsilon_{ig}
    \]
    by ordinary least squares and test the group null hypothesis $H_{0,g}: \boldsymbol{\beta}_g = \mathbf{0}$ using an F-test.
    \item If the group-level $p$-value is less than $\alpha_1$, we declare group $g$ ``active'' and proceed to variable-level screening within that group.
    \item For each active group $g$, we refit the multivariate linear model
\[
y_i = \beta_{0g} + \mathbf{x}_{i,S_g}^{\top}\boldsymbol{\beta}_g + \varepsilon_{ig},
\]
using all variables in $S_g$ simultaneously. We then test the individual null hypotheses
\[
H_{0,j}: \beta_j = 0, \quad j \in S_g,
\]
based on the corresponding $t$-statistics, and compute variable-level $p$-values.
    \item Variables with $p$-values below $\alpha_2$ are retained. Let $\mathcal{S}_1$ denote the union of all retained variables across active groups.
\end{enumerate}
The thresholds $(h, \alpha_1, \alpha_2)$ are chosen from a predefined grid via cross-validation to minimize predictive RMSE.

\noindent\textbf{Stage 3: Final penalized regression (EN / adaptive EN).}
We fit a penalized model on the screened subset $\mathcal{S}_1$:
\begin{enumerate}
    \item Let $\mathbf{X}_{\mathcal{S}_1}$ denote the $n \times |\mathcal{S}_1|$ matrix with the selected variables. We fit an elastic net model
    \[
        \hat{\boldsymbol{\beta}}^{\mathrm{EN}} 
        \;=\; \arg\min_{\boldsymbol{\beta}}
        \left\{
        \frac{1}{2n} \|\mathbf{y} - \mathbf{X}_{\mathcal{S}_1}\boldsymbol{\beta}\|_2^2 
        + \lambda \left[ \alpha \|\boldsymbol{\beta}\|_1 + \frac{1-\alpha}{2}\|\boldsymbol{\beta}\|_2^2 \right]
        \right\},
    \]
    where $\alpha = 1$ yields the LASSO and $\alpha \in (0,1)$ yields the elastic net. The penalty parameter(s) are selected by cross-validation.
    \item For the adaptive elastic net variant, we compute an initial estimate $\tilde{\boldsymbol{\beta}}$ (e.g., EN) and define adaptive weights
    \[
        w_j = \frac{1}{(|\tilde{\beta}_j| + \varepsilon)^\gamma},
    \]
    with $\gamma > 0$ and a small $\varepsilon > 0$, and refit the model with a weighted $\ell_1$ penalty $\lambda \sum_j w_j |\beta_j|$.
\end{enumerate}

We refer to Option B of this procedure as Dorfman\_sparse\_EN when using the standard EN in the final stage, and Dorfman\_sparse\_adaptive\_EN when using adaptive EN in the final stage. And we refer to Option A as the classical Dorfman\_EN in our method comparison. This entire procedure is summarized in Algorithm 1 (Appendix A). 

\subsection{Robust Spearman/OGK–Huber Dorfman–EN/Adaptive\_EN with Spearman/Sparse Robust Covariance}

To handle heavy-tailed noise, leverage points, and contamination in the dependence structure, we consider a robust extension of the Dorfman–EN framework. This method combines (i) a robust high-dimensional covariance estimator based on the OGK approach (\cite{maronna2002robust}), as implemented in (\cite{ollerer2015robust}), (ii) graphical lasso scarification, (iii) robust cluster-wise and variable-wise screening using Huber M-estimation, and (iv) a final elastic net (or adaptive elastic net) fit.

\noindent\textbf{Stage 1: Robust sparse covariance and robust grouping.} 
\textit{Option A (Spearman correlation).}
We compute the Spearman rank correlation matrix $\hat{R}^{\mathrm{sp}}$ of $\mathbf{X}$ and define the dissimilarity
\[
D_{jk} = 1 - \big| \hat{R}^{\mathrm{sp}}_{jk} \big|.
\]
Hierarchical clustering is then applied to $D$ (with dynamic tree cut) to obtain covariate groups.

\textit{Option B (OGK-based sparse correlation refinement).}
To obtain robust and sparse grouping, we combine robust covariance estimation with graphical lasso regularization:
\begin{enumerate}
    \item Apply the OGK (orthogonalized Gnanadesikan--Kettenring) estimator (\cite{maronna2002robust})
    to $\mathbf{X}$ to obtain a robust covariance matrix $\hat{\Sigma}_{\mathrm{OGK}}$.
    \item For a grid of penalty parameters $\rho \in \mathcal{R}$, apply the graphical lasso
    \[
        \hat{\Theta}_{\mathrm{rob}}(\rho)
        = \arg\min_{\Theta \succ 0}
        \big\{ \mathrm{tr}(\hat{\Sigma}_{\mathrm{OGK}}\Theta)
        - \log\det(\Theta) + \rho \|\Theta\|_1 \big\},
    \]
    yielding a robust sparse precision matrix.
    \item Invert and rescale $\hat{\Theta}_{\mathrm{rob}}(\rho)$ to obtain a sparse robust correlation matrix
    $\hat{R}^{\mathrm{sp}}_{\mathrm{rob}}(\rho)$.
    \item Form the robust dissimilarity matrix
    \[
        D^{\mathrm{rob}}_{jk}(\rho)
        = 1 - \big| \hat{R}^{\mathrm{sp}}_{\mathrm{rob},jk}(\rho) \big|.
    \]
    \item Perform hierarchical clustering with average linkage on $D^{\mathrm{rob}}(\rho)$.
    The penalty parameter $\rho^\star$ is selected by maximizing clustering purity (ARI),
    and the final groups are obtained using a dynamic tree cutting algorithm.
\end{enumerate}
These groups are designed to be robust to outliers and contaminated correlations.

\noindent\textbf{Stage 2: Robust group and variable screening (Huber).}
Given the robust groups $\{g^{\mathrm{rob}}_j\}$, we replace OLS by Huber's M-estimator to downweight outlying observations in the screening step:
\begin{enumerate}
    \item For each robust group $g$, let $S_g^{\mathrm{rob}} = \{ j : g^{\mathrm{rob}}_j = g \}$ and consider the model
    \[
        y_i = \beta_{0g} + \mathbf{x}_{i,S_g^{\mathrm{rob}}}^\top \boldsymbol{\beta}_g + \varepsilon_{ig}.
    \]
    We fit this model with Huber's M-estimator to obtain robust regression coefficients and observation weights $w_i$.
    \item Using the estimated weights, we refit a weighted least squares model and perform an F-test of $H_{0,g}: \boldsymbol{\beta}_g = \mathbf{0}$. If the robust group-level $p$-value is below a threshold $\alpha_1^{\mathrm{rob}}$, group $g$ is declared active.
    \item For each active group $g$, we refit the multivariate linear model
\[
y_i = \beta_{0g} + \mathbf{x}_{i,S_g^{\mathrm{rob}}}^{\top}\boldsymbol{\beta}_g + \varepsilon_{ig},
\]
using all variables in $S_g^{\mathrm{rob}}$ simultaneously. The model is fitted using Huber’s M-estimator to obtain robust regression coefficients and observation weights. Based on the refitted weighted least squares model, we compute variable-level robust $t$-statistics and corresponding $p$-values for each coefficient $\beta_j$, $j \in S_g^{\mathrm{rob}}$.
    \item Variables with robust $p$-values below $\alpha_2^{\mathrm{rob}}$ are retained. Let $\mathcal{S}_1^{\mathrm{rob}}$ denote the collection of all such retained variables.
\end{enumerate}
The pair $(\alpha_1^{\mathrm{rob}},\alpha_2^{\mathrm{rob}})$ is selected via cross-validation over a grid, typically allowing more liberal values than in the non-robust case to compensate for the downweighting of outliers.

\noindent\textbf{Stage 3: Final penalized regression on robustly screened variables.}
In the final step, we fit a penalized regression model on the robustly screened subset:
\begin{enumerate}
    \item Restrict the design matrix to $\mathcal{S}_1^{\mathrm{rob}}$ to obtain $\mathbf{X}_{\mathcal{S}_1^{\mathrm{rob}}}$.
    \item Fit an elastic net model
    \[
        \hat{\boldsymbol{\beta}}^{\mathrm{EN,rob}} 
        \;=\; \arg\min_{\boldsymbol{\beta}}
        \left\{
        \frac{1}{2n} \|\mathbf{y} - \mathbf{X}_{\mathcal{S}_1^{\mathrm{rob}}}\boldsymbol{\beta}\|_2^2 
        + \lambda \left[ \alpha \|\boldsymbol{\beta}\|_1 + \frac{1-\alpha}{2}\|\boldsymbol{\beta}\|_2^2 \right]
        \right\},
    \]
    with tuning parameters chosen by cross-validation.
    \item For the adaptive robust EN variant, we compute an initial estimate $\tilde{\boldsymbol{\beta}}^{\mathrm{rob}}$ and define adaptive weights
    \[
        w_j^{\mathrm{rob}} = \frac{1}{(|\tilde{\beta}^{\mathrm{rob}}_j| + \varepsilon)^\gamma},
    \]
    then refit the model with a weighted $\ell_1$ penalty $\lambda \sum_j w_j^{\mathrm{rob}} |\beta_j|$.
\end{enumerate}

We refer to Option B of this procedure as Robust OGK Dorfman-EN when using the standard EN in the final stage and as Robust OGK Dorfman-adaptive EN when using adaptive weights, built on a robust sparse covariance structure following (\cite{ollerer2015robust}), and Option A of this procedure as robust Spearman Dorfman-EN/ Adaptive-EN. The entire procedure is summarized in Algorithm~2 (Appendix A). 

\section{Numerical Studies}

\subsection{Simulation settings: normal case}

We generated $n=200$ observations with $p=Km$ predictors partitioned into $K$ non-overlapping groups of equal size $m=5$. In our experiments, we considered $K=200$ ($p=1000$). Let $G_k \subset \{1,\dots,p\}$ denote the index set for group $k$:
\[
G_k=\{(k-1)m+1,\dots,km\},\qquad k=1,\dots,K.
\]
Let $\mathbf{X}\in\mathbb{R}^{n\times p}$ be the design matrix, with $\mathbf{X}_{i,G_k}\in\mathbb{R}^m$ denoting the predictors in group $k$ for subject $i$.

\noindent\textbf{Within-group correlation structure.}
Groups were assigned one of five equicorrelation levels
\[
\rho \in \{0.1,\ 0.3,\ 0.5,\ 0.7,\ 0.9\},
\]
with an equal number of groups assigned to each level. Let $B$ denote the number of groups per correlation level (e.g., $B=20$ when $K=100$). Specifically, for $k=1,\dots,K$,
\[
\rho_k \in \{0.1,0.3,0.5,0.7,0.9\},
\]
with exactly $B$ groups assigned to each value.

For each group $k$, predictors were generated independently across subjects from a mean-zero multivariate normal distribution with an equicorrelation (compound symmetry) covariance:
\[
\mathbf{X}_{i,G_k} \sim \mathcal{N}\!\left(\mathbf{0},\,\Sigma(\rho_k)\right),\qquad
\Sigma(\rho_k) = (1-\rho_k)I_m + \rho_k \mathbf{1}_m\mathbf{1}_m^{\top}.
\]
Different groups were independent:
\[
\mathrm{Cov}\!\left(\mathbf{X}_{i,G_k},\,\mathbf{X}_{i,G_\ell}\right)=0,\qquad k\neq \ell.
\]
Equivalently, the covariance of $\mathbf{X}_i=(X_{i1},\dots,X_{ip})^\top$ is block diagonal,
\[
\Sigma=\mathrm{blockdiag}(\Sigma(\rho_1),\dots,\Sigma(\rho_K)).
\]

In implementation, $\mathbf{X}_{i,G_k}=\mathbf{Z}_{i,k} L_k$, where $\mathbf{Z}_{i,k}\sim\mathcal{N}(\mathbf{0},I_m)$ and $L_k$ is the Cholesky factor of $\Sigma(\rho_k)$.

\noindent\textbf{True signal groups and coefficients.}
Exactly one group from each correlation block was truly associated with the outcome. Let
\[
\mathcal{A} = \{1,\ B+1,\ 2B+1,\ 3B+1,\ 4B+1\}
\]
denote the set of active groups, corresponding to the first group within each correlation level.

Effect sizes were assigned by correlation block as
\[
b^{(1)}=1.0,\quad b^{(2)}=0.9,\quad b^{(3)}=0.7,\quad b^{(4)}=0.5,\quad b^{(5)}=0.3,
\]
with each active group $k\in\mathcal{A}$ receiving the coefficient associated with its block. Coefficients were constant within active groups:
\[
\beta_j =
\begin{cases}
b^{(r)}, & j\in G_k,\ k\in\mathcal{A}\ \text{and}\ k\ \text{belongs to block } r,\\
0, & \text{otherwise}.
\end{cases}
\]
Thus, all $m$ variables in each active group were signals, yielding a total of $5m=25$ nonzero coefficients.

\noindent\textbf{Outcome model and noise.}
Responses were generated according to the linear model
\[
y_i = \sum_{j=1}^{p} X_{ij}\beta_j + \varepsilon_i
= \sum_{k\in\mathcal{A}} b_k \sum_{j\in G_k} X_{ij} + \varepsilon_i,
\qquad
\varepsilon_i \stackrel{iid}{\sim} \mathcal{N}(0,1),
\]
where $b_k$ denotes the effect size associated with active group $k\in\mathcal{A}$. The noise standard deviation was fixed at $\sigma=1$. This data-generating process was repeated $B=100$ times.


\subsection{Simulation settings: robust (contaminated) case}

We considered the same grouped predictor design as in the normal case, with $n=200$ observations and $p=Km$ predictors partitioned into $K$ disjoint groups of equal size $m=5$. In our experiments, we considered both $K=200$ ($p=1000$). Group index sets are $G_k=\{(k-1)m+1,\ldots,km\}$ for $k=1,\ldots,K$. Groups were assigned one of five equicorrelation levels $\rho_k\in\{0.1,0.3,0.5,0.7,0.9\}$, with an equal number of groups per level. Let $B=K/5$ denote the number of groups per correlation block (e.g., $B=40$ when $K=200$). The clean design matrix $\mathbf{X}_{\text{clean}}\in\mathbb{R}^{n\times p}$ was generated from the Gaussian block model with block-diagonal covariance described in the normal case, and the contaminated design used for analysis is denoted by $\mathbf{X}\in\mathbb{R}^{n\times p}$.

\noindent\textbf{True signal groups and coefficients.}
Exactly one group from each correlation block was active, namely the first group within each block:
\[
\mathcal{A}=\{1,\ B+1,\ 2B+1,\ 3B+1,\ 4B+1\}.
\]
All $m=5$ variables within each active group were nonzero (so $5m=25$ active variables total). Effect sizes were assigned by block as
\[
b^{(1)}=2.0,\quad b^{(2)}=1.8,\quad b^{(3)}=1.5,\quad b^{(4)}=1.2,\quad b^{(5)}=1.0,
\]
and coefficients were constant within each active group:
\[
\beta_j=
\begin{cases}
b^{(r)}, & j\in G_k,\ k\in\mathcal{A}\ \text{and}\ k\ \text{belongs to block } r,\\
0, & \text{otherwise}.
\end{cases}
\]

\noindent\textbf{Contamination in predictors.}
Starting from $\mathbf{X}_{\text{clean}}$, we introduced multiple structured perturbations to induce outliers, spurious cross-group dependence, and batch effects, primarily in inactive groups. First, we added mild global measurement noise to all entries,
\[
\mathbf{X}\leftarrow \mathbf{X}_{\text{clean}} + \mathbf{E},\qquad E_{ij}\stackrel{iid}{\sim}\mathcal{N}(0,0.1^2).
\]
Second, we selected a small set of $n_{\text{aligned}}=8$ ``aligned'' observations $\mathcal{I}_{\text{aligned}}\subset\{1,\ldots,n\}$ and, for each $i\in\mathcal{I}_{\text{aligned}}$, replaced a random subset of $40$ inactive predictors with large shifted values (approximately centered around 8), creating high-leverage rows concentrated in inactive variables. Third, to induce false cross-group correlations among inactive predictors, we selected $|\mathcal{I}_{\text{pair}}|=20$ additional observations and, within each such row, created $n_{\text{pairs}}=20$ paired inactive variables with near-equality constraints of the form $X_{i,v_2}=X_{i,v_1}+\eta$ where $\eta\sim\mathcal{N}(0,0.1^2)$. Fourth, we introduced moderate leverage effects by selecting $|\mathcal{I}_{\text{lev}}|=12$ additional observations and inflating a subset of predictors (mostly inactive) by a multiplicative factor of 5, with an additional Gaussian perturbation. Finally, we imposed batch shifts on inactive groups by sampling two disjoint batches $\mathcal{I}_{b1}$ and $\mathcal{I}_{b2}$ of size $\lfloor 0.15n\rfloor$ each and shifting 12 inactive groups upward by $+3$ in batch 1 and 12 other inactive groups downward by $-3$ in batch 2. To avoid trivially destroying the signal, active groups received only mild symmetric batch perturbations with small mean-zero shifts.

\noindent\textbf{Outcome model with heavy tails, heteroscedasticity, and spikes.}
The response was generated from a linear signal computed using the clean design,
\[
y_i = \mathbf{x}^{\top}_{\text{clean},i}\beta + \varepsilon_i,
\]
where the error term was a sum of three components:
\[
\varepsilon_i = \varepsilon^{(t)}_i + \varepsilon^{(h)}_i + \varepsilon^{(s)}_i.
\]
The heavy-tailed component followed a scaled Student-$t$ distribution,
\[
\varepsilon^{(t)}_i = 0.8\, t_i,\qquad t_i\stackrel{iid}{\sim}t_{5},
\]
the heteroscedastic component was Gaussian with variance depending on an active-group summary (using the first active block/group),
\[
\varepsilon^{(h)}_i \sim \mathcal{N}\!\left(0,\ \sigma_i^2\right),\qquad 
\sigma_i = 0.7 + 1.0\,\big|\bar X_{\text{clean},i,G_{k^\star}}\big|,
\]
where $k^\star = 1$ (the first active group) and $\bar X_{\text{clean},i,G_{k^\star}}=\frac{1}{m}\sum_{j\in G_{k^\star}}X_{\text{clean},ij}$.
The spike component introduced large positive shocks on the aligned observations,
\[
\varepsilon^{(s)}_i=
\begin{cases}
s_i,\quad s_i\sim \mathcal{N}(8,1.5^2), & i\in \mathcal{I}_{\text{aligned}},\\
0, & \text{otherwise}.
\end{cases}
\]
We generated $R=100$ independent replicates under this contaminated mechanism.

\subsection{Competitor Methods (SIS-LASSO and Group AR2)}

\noindent\textbf{Group AR2 screening (\cite{qiu2020grouped})}

Let $Y\in\mathbb{R}^n$ be the response and let $X\in\mathbb{R}^{n\times p}$ be the predictor matrix.
Assume the $p$ predictors are partitioned into $K$ non-overlapping groups
$\{X_1,\dots,X_K\}$, where $X_j\in\mathbb{R}^{n\times p_j}$ denotes the submatrix
corresponding to group $j$ and $\sum_{j=1}^K p_j=p$.
For each group $j$, Group AR2 fits a least-squares regression of $Y$ on $X_j$ and computes a
group-level screening score based on the adjusted coefficient of determination.

\noindent\textbf{Group-wise regression and raw $R_j^2$.}

For group $j$, the fitted values are
\[
\widehat{Y}_j \;=\; X_j\widehat{\beta}_j
\;=\; X_j (X_j^\top X_j)^{-1}X_j^\top Y,
\]
and the raw group $R^2$ is defined as
\[
R_j^2 \;=\; 1-\frac{\|Y-\widehat{Y}_j\|_2^2}{\|Y-\bar{Y}\mathbf{1}\|_2^2},
\qquad
\bar{Y}=\frac{1}{n}\mathbf{1}^\top Y,
\]
where $\mathbf{1}$ is the $n$-vector of ones.

\noindent\textbf{Adjusted $R_j^2$ (group score).}
To correct for group size $p_j$, the adjusted $R^2$ score is
\[
\bar{R}_j^2
\;=\;
\frac{n-1}{n-p_j-1}R_j^2 - \frac{p_j}{n-p_j-1}.
\]
Group AR2 uses $\bar{R}_j^2$ as the group screening statistic.

\noindent\textbf{Selection rule and definition of $d$.}
Let $d$ denote the \emph{number of groups retained} after screening.
Rank groups by $\bar{R}_j^2$ in descending order and select the top $d$ groups:
\[
\mathcal{M}^{g,\mathrm{AR2}}_{d}
\;=\;
\Big\{ j\in\{1,\dots,K\}:\ \bar{R}_j^2 \text{ is among the largest } d \text{ values} \Big\}.
\]
Equivalently, the screened predictor set is given by
$\bigcup_{j\in\mathcal{M}^{g,\mathrm{AR2}}_{d}} \{\text{columns of }X_j\}$.

$d$ is chosen by the common SIS rule: $d = min\{K, \lfloor n/log(n)\rfloor\}$
(\cite{qiu2020grouped}), with prespecified $K = 100$ groups when $p = 500$, and $K = 200$ groups when $p = 1000$.

\noindent\textbf{SIS-LASSO screening (\cite{fan2008sure})}

Sure independence screening (SIS) ranks predictors by their marginal utility with respect to the response.
For each predictor $X_j\in\mathbb{R}^n$, define the marginal correlation score
\[
\omega_j = \left|\mathrm{cor}(X_j, Y)\right|
= \left|\frac{\sum_{i=1}^n (X_{ij}-\bar X_j)(Y_i-\bar Y)}
{\sqrt{\sum_{i=1}^n (X_{ij}-\bar X_j)^2}\sqrt{\sum_{i=1}^n (Y_i-\bar Y)^2}}\right|.
\]
\begin{sloppypar}
Let $\omega_{(1)}\ge \omega_{(2)}\ge \cdots \ge \omega_{(p)}$ denote the ordered scores. SIS retains the top $d$ predictors,
\end{sloppypar}
\[
\mathcal{M}^{\mathrm{SIS}}_{d}
=
\left\{ j\in\{1,\dots,p\}:\ \omega_j \text{ is among the largest } d \right\}.
\]
SIS-LASSO then fits a LASSO regression on the reduced set of predictors $X_{\mathcal{M}^{\mathrm{SIS}}_{d}}$:
\[
\widehat{\beta}^{\mathrm{SIS\text{-}LASSO}}
=
\arg\min_{\beta\in\mathbb{R}^{d}}
\left\{
\frac{1}{2n}\|Y-X_{\mathcal{M}^{\mathrm{SIS}}_{d}}\beta\|_2^2
+
\lambda \|\beta\|_1
\right\},
\]
where $\lambda>0$ is typically chosen by cross-validation or an information criterion. The final selected
variables are those with nonzero estimated coefficients.

\begin{table}[htbp]
\small
\centering
\caption{Performance metrics of methods under the normal scenario ($p=1000$). Results are reported as mean (sd).}
\begin{tabular}{lccccc}
\hline
Method & Variables & TPR & FDR & F1 & RMSE \\
\hline
Dorfman\_sparse\_Adaptive\_EN 
& 24.9 (5.8) 
& 0.924 (0.078) 
& 0.070 (0.019) 
& 0.926 (0.129) 
& 1.256 (0.278) \\

Robust\_OGK\_Dorfman\_Adaptive\_EN 
& 24.5 (8.7) 
& 0.916 (0.042) 
& 0.063 (0.027) 
& 0.925 (0.105) 
& 1.641 (0.389) \\

Adaptive\_EN 
& 26.3 (3.2) 
& 0.867 (0.028) 
& 0.169 (0.080) 
& 0.846 (0.045) 
& 1.298 (0.095) \\

Dorfman\_EN 
& 39.9 (5.6) 
& 0.916 (0.073) 
& 0.413 (0.112) 
& 0.713 (0.103) 
& 1.931 (0.593) \\

Group\_AR2\_gp\_LASSO 
& 68.2 (15.6) 
& 0.992 (0.056) 
& 0.585 (0.115) 
& 0.748 (0.061) 
& 7.553 (1.25) \\

Robust\_OGK\_Dorfman\_EN 
& 28.1 (5.1) 
& 0.791 (0.131) 
& 0.288 (0.097) 
& 0.743 (0.090) 
& 2.446 (0.910) \\

SIS\_LASSO 
& 16.0 (0) 
& 0.578 (0.057) 
& 0.097 (0.089) 
& 0.705 (0.069) 
& 3.137 (0.497) \\

EN 
& 55.4 (10.6) 
& 0.982 (0.021) 
& 0.542 (0.096) 
& 0.620 (0.085) 
& 1.574 (0.239) \\
\hline
\end{tabular}
\end{table}

\noindent\textbf{Normal scenario.}
In the normal (uncontaminated) simulation setting, the proposed Dorfman-based screening procedures achieved the strongest overall performance by balancing sensitivity and false discovery control while selecting a moderate number of variables. In particular, \textbf{Dorfman\_sparse\_Adaptive\_EN} produced the best results, with the highest average \textbf{$F_1 = 0.926$}, strong sensitivity (\textbf{TPR $= 0.924$}), and a low false discovery rate (\textbf{FDR $= 0.070$}), while selecting only \textbf{24.9 variables} on average, which is closest to the true value of 25, also achieving the highest predictive power (lowest RMSE $= 1.256$). On the other hand, we observed a trend of robust methods underperforming their non-robust counterparts, namely \textbf{Robust\_Sparse\_Dorfman\_Adaptive\_EN vs.\ Sparse\_Dorfman\_Adaptive\_EN ($F_1$: 0.925 vs.\ 0.926; RMSE: 1.641 vs.\ 1.256)}, which is expected under normal conditions when linear assumptions hold. In contrast, adaptive EN showed competitive but weaker performance (0.846 vs.\ 0.926) and substantially higher false discovery rates (0.169), which is approximately twice that of the sparse adaptive Dorfman procedure. The classical Dorfman approach (i.e., Dorfman\_EN) underperformed: although it maintained high sensitivity (\textbf{TPR: 0.916}), its higher false discovery rate (\textbf{0.413}) reduced overall performance (\textbf{$F_1$: 0.713, RMSE: 1.93}).

\textbf{Group\_AR2} achieved near-perfect sensitivity (\textbf{TPR $= 0.992$}) but selected far more variables (\textbf{68.2}) and produced a high false discovery rate (\textbf{0.585}), suggesting that it tends to over-select and introduce many false positives. This occurs because group\_AR2\_gp\_LASSO selects all variables across all positive groups in the first stage, and the second stage applies a group LASSO to these variables (\cite{qiu2020grouped}), which itself has a relatively high false-positive rate (\cite{simon2013sparse}). \textit{The finding that our Dorfman method outperforms gAR2 under independent groups in the within-group correlation scenario (one of six scenarios presented in (\cite{qiu2020grouped})), outperforming group-SIS and group-HOLP.}

Similarly, vanilla \textbf{EN} selected very large models (\textbf{55.4 variables}) with high false discovery rates (both approximately \textbf{0.542}), resulting in the lowest $F_1$ score (\textbf{0.620}). Finally, \textbf{SIS\_LASSO} achieved the lowest FDR (\textbf{0.097}), but its lower sensitivity (\textbf{TPR: 0.578}) limited overall recovery performance (\textbf{$F_1: 0.705$}). As noted, the LASSO penalty constrains the number of variables selected. Although performance is expected to improve as $n$ increases, in ultra-high-dimensional settings where $p \gg n$, performance remains limited.

Overall, these results show that the adaptive sparse Dorfman procedure provides the best trade-off between selecting true signals and controlling false positives in the clean high-dimensional setting, and that it outperforms not only competing methods, but also the robust variant.

\begin{table}[htbp]
\small
\centering
\caption{Performance metrics of methods under the robust (contaminated) scenario ($p=1000$). Results are reported as mean (sd).}
\begin{tabular}{lccccc}
\hline
Method & Variables & TPR & FDR & F1 & RMSE \\
\hline
Robust\_OGK\_Dorfman\_Adaptive\_EN
& 30.5 (7.6)
& 0.896 (0.050)
& 0.261 (0.113)
& 0.809 (0.078)
& 5.249 (1.081) \\

Robust\_OGK\_Dorfman\_EN
& 29.9 (8.0)
& 0.884 (0.049)
& 0.253 (0.118)
& 0.807 (0.084)
& 5.731 (1.662) \\

Dorfman\_sparse\_Adaptive\_EN
& 22.7 (4.5)
& 0.647 (0.198)
& 0.280 (0.095)
& 0.676 (0.182)
& 7.368 (2.614) \\

Dorfman\_EN
& 35.5 (3.1)
& 0.860 (0.090)
& 0.389 (0.095)
& 0.713 (0.092)
& 5.783 (1.519) \\

group\_AR2\_gp\_LASSO
& 72.4 (14.9)
& 0.987 (0.061)
& 0.612 (0.102)
& 0.741 (0.061)
& 7.864 (1.314) \\

SIS\_LASSO
& 16 (0)
& 0.613 (0.041)
& 0.042 (0.065)
& 0.748 (0.050)
& 7.503 (1.152) \\

Adaptive\_EN
& 31.7 (2.7)
& 0.880 (0.025)
& 0.302 (0.047)
& 0.777 (0.024)
& 5.325 (0.809) \\

EN
& 42.0 (12.7)
& 0.887 (0.053)
& 0.439 (0.136)
& 0.677 (0.100)
& 5.305 (0.758) \\
\hline
\end{tabular}
\end{table}

\begin{table}[htbp]
\small
\centering
\caption{Runtime statistics (mean (sd)) for top-performing Dorfman methods in comparison to competitors (seconds).}
\begin{tabular}{lcc}
\hline
Method & Normal & Robust \\
\hline
Robust\_OGK\_Dorfman\_Adaptive\_EN
& 7.743 (0.326)
& 8.293 (0.956) \\

Dorfman\_sparse\_Adaptive\_EN
& 105.702 (1.407)
& 118.876 (2.235) \\

SIS\_LASSO
& 0.386 (0.183)
& 0.331 (0.177) \\

Adaptive\_EN
& 0.025 (0.001)
& 0.029 (0.008) \\

EN
& 0.016 (0.003)
& 0.018 (0.006) \\
\hline
\end{tabular}
\end{table}

\noindent\textbf{Corrupted scenario.}
Under the contaminated (robust) scenario (Table~2), the proposed robust Dorfman variants achieved the strongest overall screening performance, yielding the highest average $F_1$ scores while maintaining relatively low false discovery. In particular, \textbf{Robust\_OGK\_Dorfman\_Adaptive\_EN} achieved the best overall balance, with \textbf{Avg\_$F_1 = 0.809$}, \textbf{Avg\_TPR $= 0.896$}, and \textbf{Avg\_FDR $= 0.261$}. It also achieved the lowest RMSE (5.249), corresponding to the strongest predictive performance. The \textbf{Robust\_OGK\_Dorfman\_EN} method performed similarly well (\textbf{Avg\_$F_1 = 0.807$}, \textbf{Avg\_TPR $= 0.884$}, \textbf{Avg\_FDR $= 0.253$}, and \textbf{Avg\_Vars $= 29.9$} versus 30.5 in the best-performing method). In this scenario, the robust methods outperformed classical Dorfman approaches (\texttt{Dorfman\_sparse\_Adaptive\_EN} and \texttt{Dorfman\_EN}), as expected under contamination, demonstrating the reliability of the robust variants.

Interestingly, the sparse adaptive Dorfman method, which performed best in the normal case, underperformed even relative to \texttt{Dorfman\_EN} in the robust setting ($F_1$: 0.676 vs.\ 0.713; RMSE: 7.368 vs.\ 5.783). This further highlights that graphical LASSO-derived sparse precision matrices can become statistically unstable under model misspecification in contaminated settings, underscoring the utility of the OGK estimator in providing contamination-resistant covariance estimates.

As seen in the normal case, Group\_AR2\_gp\_LASSO produced the \textbf{largest selected model} (\textbf{Avg\_Vars $= 72.4$}) and achieved the \textbf{highest Avg\_TPR $= 0.987$}, but suffered from substantially inflated false discovery (\textbf{Avg\_FDR $= 0.612$}) and the worst predictive performance (\textbf{Avg\_RMSE $= 7.864$}). This behavior is expected, particularly when contamination induces spurious cross-group dependence or when imperfect group definitions and outliers distort group-level screening, leading to high FDR despite strong sensitivity. \textbf{SIS\_LASSO} was the most conservative method, selecting the fewest variables (\textbf{Avg\_Vars $= 16.0$}) due to the imposed screening constraint on $d$, and achieved the lowest false discovery (\textbf{Avg\_FDR $= 0.042$}), but missed a substantial fraction of true signals (\textbf{Avg\_TPR $= 0.613$}) and exhibited relatively poor predictive accuracy (\textbf{Avg\_RMSE $= 7.503$}). Similarly, vanilla \textbf{EN} selected much larger models (\textbf{Avg\_Vars $= 42$}) with high sensitivity (\textbf{Avg\_TPR $= 0.887$}) but elevated false discovery (\textbf{Avg\_FDR $= 0.439$}), resulting in lower overall performance (\textbf{Avg\_$F_1 = 0.677$}).

\noindent\textbf{Runtime comparison.}
The average runtime of \textbf{Robust\_OGK\_Dorfman\_Adaptive\_EN} on a MacBook Pro M4 using RStudio version 2026.01.0-392 is 7.743 seconds in the normal case and 8.293 seconds in the robust case. Although the runtime is longer than that of competitor methods such as SIS-LASSO, adaptive EN, and vanilla EN, the improvements in variable selection accuracy ($F_1$) and predictive performance (RMSE) still make it a superior choice. In contrast, Dorfman's sparse adaptive EN, although slightly outperforming robust methods under normal conditions, performs poorly under robust conditions across both selection accuracy and prediction. Moreover, its runtime is 105.702 seconds in the normal case and 118.876 seconds in the robust case, which is more than 10 times slower than the robust variants.

These runtime differences arise primarily from covariance estimation and hyperparameter search. In the normal pipeline, empirical covariance estimation followed by graphical LASSO and clustering has complexity $O(p^3)$ (\cite{MazumderHastie2012EJS}), whereas OGK relies on univariate robust estimates with complexity approximately $O(np^2)$ and often improves numerical conditioning, accelerating convergence (\cite{ollerer2015robust}). Additionally, the standard pipeline performs hierarchical clustering with a grid search over cut height and testing thresholds, yielding numerous cross-validation combinations, whereas dynamic tree-cut clustering eliminates the need for cross-validation over cut heights, thereby substantially reducing search complexity.

Overall, results across variable selection metrics ($F_1$, TPR, FDR), number of selected variables, predictive RMSE, and runtime demonstrate that the robust OGK-based adaptive Dorfman method \textbf{is the method of choice} for high-dimensional settings ($p>n$). Especially when data are noisy or contaminated, it performs comparably to sparse adaptive Dorfman under ideal normal assumptions, which rarely hold in practice. Real-world biomedical data (e.g., genomics, multi-omics, and EHR data) are typically noisy and contaminated, and we will demonstrate practical performance using IC50 prediction with GDSC data in Section~4.

We further expect a key advantage of the robust Dorfman framework to be its scalability to ultra-high-dimensional regimes. Under scenarios where $p \gg n$ (e.g., $p = 100{,}000$, $n = 200$), algorithmic and hyperparameter-selection modifications are required (see \textbf{Limitations and Future Directions} and Appendix~B). Classical EN and adaptive EN rely on low-dimensional asymptotics and the irrepresentable condition, and break down in ultra-high dimensions due to noise accumulation (\cite{fan2008high}), violation of selection conditions~\cite{zhao2006model}, and miscalibrated adaptive weights (\cite{zou2006adaptive}). Even robust alternatives such as Adaptive PENSE (\cite{kepplinger2023robust}) lose breakdown and oracle properties when $p \gg n$. In contrast, Dorfman-based methods are hierarchical and screening-driven, reducing dimensionality prior to penalization while exploiting group sparsity. This structure naturally supports blockwise computation and scalable hyperparameter optimization. We therefore hypothesize that, once scaled, Dorfman-based approaches will outperform EN, adaptive EN, and robust penalized alternatives by an even larger margin in ultra-high-dimensional regimes, where competing methods become unstable while our framework remains computationally and statistically feasible.

\section{Application to the GDSC NSCLC trametinib dataset}
We applied the proposed Dorfman screening framework to the Genomics of Drug Sensitivity in Cancer (GDSC) dataset, focusing on non-small cell lung cancer (NSCLC) cell lines treated with the MEK inhibitor trametinib. After filtering for NSCLC histology, the dataset comprises 121 cell lines with gene expression profiles measured across 17,736 genes and corresponding trametinib IC50 values. Genes with negligible expression across NSCLC cell lines were first removed to reduce technical noise and ensure model stability. Specifically, firstly, genes with low expression levels ($\log_2(\mathrm{TPM}+1) < 1$) in more than $80\%$ of NSCLC cell lines were excluded. Among the remaining genes, variability across NSCLC cell lines was quantified using the median absolute deviation (MAD), and the top 1,000 most variable genes were retained for subsequent Dorfman screening and penalized regression analyses. These genes were then mapped to 50 Hallmark pathways, ensuring coverage of core oncogenic and signaling processes. Genes in the filtered GDSC dataset were first clustered into 50 bigger hallmark groups. Hierarchical clustering/Dynamic HC is then applied to all groups with genes $\ge 3$ (only one group has 2 genes, so that group is not further clustered). The rest of Dorfman procedure was run subsequently (Figure~2), and the selected genes of each method is compared against a set of signature genes based on NSCLC specific clinical evidence (Figure~2), and methods were compared for the proportion of clinical hits (tiers 1 and 2) (number of clinical hits$/\,$total number of signature genes in tier 1 and tier 2), the number of genes in each tier, as well as prediction RMSE.

\begin{figure}[htbp]
    \centering
    \includegraphics[width=0.9\linewidth]{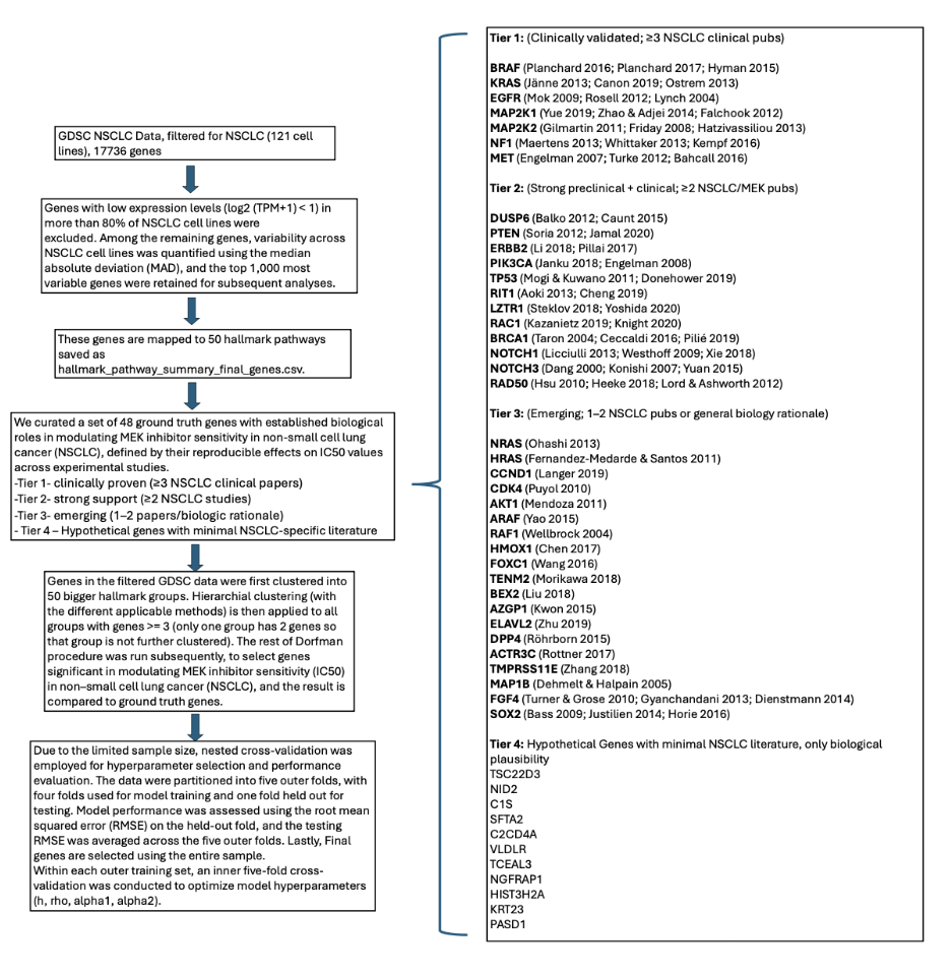}
    \caption{Workflow for selecting and validating genes selected by Dorfman and other competitor methods.}
    \label{fig:2}
\end{figure}

To benchmark model discoveries against established biology, we organized genes into tiered signature sets reflecting the strength of existing evidence linking them to MEK inhibition in NSCLC. Tier 1 (Clinically-Validated) genes were defined as those supported by three or more high-impact, NSCLC-specific clinical publications that directly link them to \textbf{MEK/MAPK response or resistance (e.g., BRAF, KRAS, MAP2K1/2, NF1, MET)}. Tier 2 (Strong Preclinical/Clinical) genes were supported by at least two solid studies in NSCLC or closely related MEK-inhibitor contexts (e.g., \textit{DUSP6, PIK3CA, PTEN, ERBB2}) but not yet fully clinically established. Tier 3 (Emerging Evidence) included genes with limited but suggestive evidence (one to two publications or a strong mechanistic rationale), whereas the final tier encompassed novel or hypothetical genes with minimal prior NSCLC-specific literature. This tiered structure enabled systematic evaluation of whether data-driven selections recovered known biology and identified plausible novel contributors to trametinib response. This resulted in 7 genes in tier 1, 11 in tier 2, 19 in tier 3, and 11 in tier 4. We ensure that the selected genes for each Dorfman method are included in the gene set, grouped by their corresponding evidence.

Due to the limited sample size, nested cross-validation was employed for hyperparameter selection and performance evaluation. The data were partitioned into five outer folds, with four folds used for model training and one fold held out for testing. Model performance was assessed using the root mean squared error (RMSE) on the held-out fold, and the testing RMSE was averaged across the five outer folds. Lastly, genes are selected from the entire sample using the median optimal parameter values determined for each fold. Within each outer training set, an inner five-fold cross-validation was conducted to optimize model hyperparameters $(h, \rho, \alpha_1, \alpha_2)$ using our developed methods.

\begin{table}[htbp]
\small
\centering
\caption{Comparison of each method on final genes, clinical significance, and prediction RMSE.}
\setlength{\tabcolsep}{3pt}
\begin{tabular}{l c c p{5.2cm} c c c c}
\hline
Method & RMSE & $N_{\text{Genes}}$ & Genes & T1 & T2 & T3 & Hyp. \\
\hline

\multirow{2}{*}{\shortstack[l]{Robust\_Dorfman\\Adaptive\_EN (OGK)}}
& \multirow{2}{*}{2.17}
& \multirow{2}{*}{12}
& BRAF, DUSP6, AZGP1, FGF4, TSC22D3, ELAVL2, DPP4, HMOX1,
& \multirow{2}{*}{1} & \multirow{2}{*}{1} & \multirow{2}{*}{6} & \multirow{2}{*}{4} \\
&&& FOXC1, NID2, HIST3H2A, C2CD4A &&&& \\

\multirow{2}{*}{\shortstack[l]{Robust\_Dorfman\\EN (OGK)}}
& \multirow{2}{*}{2.33}
& \multirow{2}{*}{10}
& BRAF, KRAS, DUSP6, BEX2, FGF4,
& \multirow{2}{*}{2} & \multirow{2}{*}{1} & \multirow{2}{*}{6} & \multirow{2}{*}{1} \\
&&& DPP4, HMOX1, HIST3H2A, MAP1B, TENM2 &&&& \\

Dorfman\_sparse\_Adaptive\_EN
& 2.38 & 3
& KRAS, DUSP6, BEX2
& 1 & 1 & 1 & 0 \\

\multirow{2}{*}{Dorfman\_sparse\_EN}
& \multirow{2}{*}{2.24}
& \multirow{2}{*}{10}
& BRAF, MAP2K1, C1S, BEX2, MAP1B,
& \multirow{2}{*}{2} & \multirow{2}{*}{0} & \multirow{2}{*}{4} & \multirow{2}{*}{4} \\
&&& SFTA2, C2CD4A, AZGP1, TMPRSS11E, VLDLR &&&& \\

\multirow{2}{*}{Dorfman\_EN}
& \multirow{2}{*}{2.43}
& \multirow{2}{*}{7}
& MET, BEX2, DPP4, TENM2,
& \multirow{2}{*}{1} & \multirow{2}{*}{0} & \multirow{2}{*}{6} & \multirow{2}{*}{0} \\
&&& TMPRSS11E, SOX2, FOXC1 &&&& \\

Adaptive\_LASSO
& 2.46 & 3
& BEX2, DUSP6, FGF4
& 0 & 1 & 2 & 0 \\

Adaptive\_EN
& 2.41 & 5
& BEX2, DUSP6, TENM2, FGF4, FOXC1
& 0 & 1 & 4 & 0 \\

LASSO
& 2.49 & 7
& BEX2, DUSP6, DPP4, TENM2, TMPRSS11E, ACTR3C, FOXC1
& 0 & 1 & 6 & 0 \\

EN
& 2.51 & 5
& BEX2, DUSP6, ACTR3C, FOXC1, TENM2
& 0 & 1 & 4 & 0 \\

\multirow{3}{*}{SIS\_LASSO}
& \multirow{3}{*}{2.84}
& \multirow{3}{*}{18}
& ELAVL2, BEX2, NID2, DUSP6, KRT23, DPP4,
& \multirow{3}{*}{0} & \multirow{3}{*}{1} & \multirow{3}{*}{10} & \multirow{3}{*}{7} \\
&&& MAP1B, SFTA2, PASD1, NGFRAP1, AZGP1, TENM2, \\
&&& HMOX1, TCEAL3, TMPRSS11E, ACTR3C, VLDLR, FOXC1 &&&& \\

\hline
\end{tabular}
\end{table}

Clinical relevance ratio $\textbf{Clinical\_Ratio\_GT} = \frac{\text{Tier1\_Hits + Tier2\_Hits}}{|\text{Tier1\_GT}| + |\text{Tier2\_GT}|}$ measures how much of the known clinically relevant ground truth (Tier1+2) each method recovers. 

\begin{figure}
    \centering
    \includegraphics[width=1\linewidth]{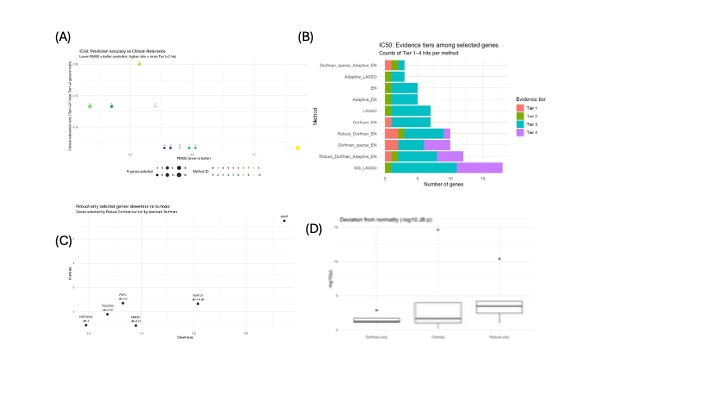}
    \caption{Features of genes selected by robust Dorfman methods in comparison to classical Dorfman methods, and in comparison to other methods. 
    (A) \textbf{IC50 prediction accuracy vs clinical relevance across gene selection methods.} Each point represents one method, plotted by RMSE (x-axis; lower is better) and the clinical relevance ratio (y-axis; Tier1+Tier2 hits divided by total Tier1+Tier2 ground truth). Point size indicates the number of selected genes. \textbf{Method IDs:} 1 = Adaptive\_EN, 2 = Adaptive\_LASSO, 3 = Dorfman\_EN, 4 = Dorfman\_sparse\_Adaptive\_EN, 5 = Dorfman\_sparse\_EN, 6 = EN, 7 = LASSO, 8 = Robust\_Dorfman\_Adaptive\_EN, 9 = Robust\_Dorfman\_EN, 10 = SIS\_LASSO. Overall, \textbf{Robust\_Dorfman\_EN (ID 9)} provides the strongest clinical enrichment and good predictive accuracy, while \textbf{Robust\_Dorfman\_Adaptive\_EN (ID 8)} achieves the best RMSE with strong clinical relevance. While Sparse Dorfman methods (Dorfman\_sparse\_Adaptive\_EN and Dorfman\_sparse\_EN) achieved the same clinical ratio as \textbf{Robust\_Dorfman\_Adaptive\_EN}, they had higher RMSE and lower predictive accuracy. In contrast, RMSE was higher in vanilla EN and LASSO, with SIS-LASSO the worst performer.
    (B) Bar plot of evidence tiers the genes selected by each method fall into. Red shows tier 1 clinical genes. Dorfman methods consistently select tier 1 genes that other methods fail to select; robust Dorfman EN and Dorfman Sparse EN select the most tier 1 genes, whereas adaptive LASSO/EN and vanilla LASSO/EN select only tier 2 and 3 genes. In contrast, SIS-LASSO selects the greatest number of genes, but also the most number of tier 4 genes with no prior literature.
    (C) Skewness and kurtosis, along with JB score for genes selected by robust Dorfman methods but missed by traditional Dorfman. JB score is calculated from $-\log_{10}(\text{JB p-value})$, where JB statistic $= (n/6)\left(S^2 + ((K-3)^2)/4\right)$, $n =$ sample size, $S =$ sample skewness, $K =$ sample kurtosis (\textbf{not excess kurtosis}), where 3 is the kurtosis of a normal distribution. Under $H_0$, $\text{JB} \sim \chi^2(2)$. Robust methods identified genes with extreme deviations from normality, e.g., NID (JB $\sim \infty$, p-value $\sim 0$) and ELAV2 (JB = 10.38), as indicated by higher skewness and kurtosis.
    (D) JB score ($-\log(p\text{-value})$) of selected genes by Dorfman-only, overlap, and robust-only. There is clearly greater deviation from normality for robust-only selected genes than for Dorfman-only \textbf{selected genes, confirming the robustness of our robust-Dorfman methods in identifying informative but non-normal genes.}}
    \label{fig:3}
\end{figure}

Upon examining our IC50 data, we observed several deviations from normality: non-normal IC50 distributions, outliers, heteroscedasticity, and skewness and kurtosis in gene expression (Appendix A, Fig.~4). We employed our top-performing methods in both our normal and corrupt scenarios, namely, the robust methods \texttt{Robust\_Dorfman\_Adaptive\_EN} and robust \texttt{Dorfman\_EN} with sparse correlation derived from the OGK estimator, classical methods \texttt{Dorfman\_sparse\_Adaptive\_EN} and \texttt{Dorfman\_sparse\_EN} using sparse correlation from graphical LASSO, as well as the traditional \texttt{Dorfman\_EN} with Pearson correlation (for comparison). The methods we are comparing are adaptive EN/LASSO, vanilla EN/LASSO, and SIS-LASSO.

The final genes selected, clinical significance, and prediction RMSE are shown in Table~4. Across all methods, genes repeatedly selected were enriched for known modulators of MEK inhibitor sensitivity; however, clear differences emerged in both predictive performance and biological interpretability. The robust Dorfman methods achieved the lowest prediction error, with Robust Dorfman–Adaptive Elastic Net yielding the best RMSE (2.17) while selecting a compact set of genes that included multiple Tier~1 and Tier~2 signatures such as \textit{BRAF, KRAS, DUSP6}, alongside emerging candidates (\textit{HMOX1, FOXC1, AZGP1}), while achieving the second-best clinical ratio, whereas \textbf{Robust\_Dorfman\_EN (ID 9) is the clear winner as it} provides the strongest clinical enrichment with good prediction accuracy, recovering the same signature genes \textit{BRAF, KRAS, DUSP6} as well as similar emerging genes (\textit{HMOX1, FGF4, MAP1B}).

While classical sparse Dorfman methods (\texttt{Dorfman\_sparse\_Adaptive\_EN} and \texttt{Dorfman\_sparse\_EN}) achieved the same clinical ratio as \textbf{Robust\_Dorfman\_Adaptive\_EN}, they had higher RMSE and lower predictive accuracy. This demonstrates the performance of robust methods under violations of linear assumptions inherent in the data (non-normal IC50, outliers, heteroscedasticity, skewness, and kurtosis in gene expression). Notably, robust methods exclusively identified genes with extreme deviation from normality, e.g., NID (JB $\sim \infty$, meaning p-value $\sim 0$), and ELAV2 (JB $= 10.38$), also shown by both higher skewness and kurtosis (Fig.~3C), \textit{and have a higher JB score than genes selected exclusively by classical Dorfman methods} (Fig.~3D). Nevertheless, both robust and classical Dorfman models consistently recovered clinically validated Tier~1 drivers (\textit{BRAF, MAP2K1, KRAS, MET1}) with stable inclusion across folds (as in Tier~1 ratio and clinical hits, Fig.~3B), which were missed by traditional methods. For example, Robust Dorfman-EN recovered both BRAF and KRAS; Dorfman-sparse adaptive EN recovered KRAS; Dorfman-sparse EN recovered both KRAS and MAP2K1; and Dorfman-EN identified MET as a Tier~1 gene. In general, Dorfman methods have lower RMSE than the rest (2.17--2.43), compared to other methods (2.41--2.84), with SIS-LASSO having the worst prediction (RMSE = 2.84). Notably, adaptive LASSO and adaptive EN still perform better than vanilla LASSO, EN, and SIS-LASSO (2.41--2.46 vs.\ 2.49--2.84) in prediction ability due to pathway structure, but fail to excel in clinical ratio.

The two key regulators consistently identified by the Dorfman-based methods, BRAF and KRAS, exhibit well-characterized and biologically meaningful patterns of trametinib sensitivity in NSCLC. In the GDSC dataset, BRAF V600E--mutant cell lines display substantially lower trametinib IC50 values (\cite{Iorio2016PharmacoGenomics}). In contrast, BRAF wild-type NSCLC cell lines show higher and more heterogeneous IC50 values, consistent with variable pathway activation and the presence of alternative resistance mechanisms (\cite{Prahallad2012BRAFColonEGFR}). Similarly, KRAS-mutant cell lines exhibit intermediate, context-dependent sensitivity to trametinib (\cite{Chen2020KRASMEK}), consistent with prior clinical and preclinical evidence that MEK inhibition alone yields heterogeneous responses in KRAS-driven NSCLC (\cite{Janne2013SelumetinibDocetaxel}). MAP2K1, uniquely identified by Dorfman-sparse EN, is crucial because it is a direct target of MEKi (\cite{Flaherty2012BRAFMek}). \textbf{MET was uniquely identified by Dorfman\_EN}, highlighting that even the non-robust Dorfman framework can recover \textbf{biologically meaningful NSCLC drivers}, since \textbf{MET signaling is a known resistance and bypass pathway that can modulate response to MEK inhibition} (\cite{Engelman2007METERBB3,Turke2010METAmplification}).

Non-Dorfman methods (adaptive EN/LASSO and vanilla EN/LASSO) tended to select broader, more heterogeneous gene sets, including Tier~2 and Tier~3 sets, with limited enrichment of clinically validated MEK pathway components and the omission of Tier~1 genes entirely (Fig.~3B). Although some key regulators, such as \textit{DUSP6} and \textit{BEX2}, appeared across multiple approaches, these models exhibited higher RMSE and less alignment with established MEK biology. SIS-LASSO, on the other hand, selected the largest number of genes and also the largest number of Tier~4 genes, none of which had prior literature in NSCLC or general biology. Our results demonstrate that incorporating structured grouping and robust correlation estimation substantially improves both predictive accuracy and biological relevance. The superior performance of robust Dorfman methods is consistent with simulation findings, further validating them as the best choice.

\subsection{Limitations and Future Directions}

\noindent\textbf{Extension to ultra-high dimensions.}
When extending to ultra-high dimensions ($p \gg n$, e.g., $p = 100{,}000$), several algorithmic changes must be made. Full Spearman or Pearson correlation matrices become infeasible, and the first step must be replaced by marginal pre-screening based on top correlations (or Kendall correlations in the robust case) with $Y$ (\cite{fan2008high}). Retained variables are then partitioned into blocks. Subsequently, the distributed block adaptive graphical LASSO is applied to empirical covariance (normal case) (\cite{witten2011new}), or a blockwise OGK estimator is used (robust case) (\cite{maronna2002robust}), followed by hierarchical clustering or dynamic tree cutting within each block. Groups derived within blocks can then be merged across blocks using locality-sensitive hashing (LSH) (\cite{andoni2008near}), substantially reducing pairwise comparisons. The resulting groups are then used in stages: group testing, individual testing, and penalized regression. Details appear in Appendix~B, Section~B.4.

\noindent\textbf{Parameter optimization in ultra-high dimensions (e.g., $p=100{,}000$).}
The graphical LASSO penalty parameter, previously optimized using Adjusted Rand Index (ARI) when groups are known, must instead be chosen by maximizing average ARI across blocks (\cite{friedman2008sparse}). Exhaustive grid search is computationally prohibitive and unstable due to discrete threshold jumps. To overcome this, we propose \textbf{Scalable Hyperparameter Optimization for Ultra-High Dimensions} based on Bayesian optimization (\cite{snoek2012practical,frazier2018tutorial}), which constructs a Gaussian-process surrogate and samples promising hyperparameter configurations, requiring 10--50$\times$ fewer evaluations than grid search while converging to near-optimal solutions (Appendix~B, Sections~B.1--B.3).

\noindent\textbf{Time complexity analysis of Dorfman screening.}
The classical algorithm is dominated by initial screening and grouping. For $p=100{,}000$, $n=200$, marginal screening costs $O(np)=O(2\times10^7)$. Blockwise graphical LASSO over $K=40$ blocks of size $\approx100$ yields roughly $O(8\times10^6)$ operations. LSH merging reduces comparisons to about $O(10^3)$, giving overall complexity $\approx O(3\times10^7)$, feasible on moderate HPC systems.

The robust variant is more computationally intensive because it employs robust estimators. Kendall’s $\tau$ screening is $O(nplog n)\approx O(10^7\log n)$ after optimization. Distributed OGK-based covariance adds approximately $O(6\times10^8)$ operations, with merging and consensus steps raising total cost to $\sim O(10^9)$ operations.

\noindent\textbf{Feasibility with HPC.}
Both algorithms are feasible on modern HPC systems. The classical version runs on a multi-core node with 32-64\,GB RAM within minutes to hours. The robust version benefits from distributed parallelization and GPU acceleration, using MPI or Spark and optimized linear algebra libraries, thereby reducing runtime to a few hours. Memory requirements remain manageable due to blockwise processing and sparse representations. Thus, although computationally demanding, the robust approach remains practical for $p=100{,}000$ and $n=200$ in HPC environments.

\section{Conclusion}

Our results demonstrate that robust-OGK-Dorfman-Adaptive EN/EN is the method of choice for feature selection in high-dimensional, group-correlated data. In simulations under both normal and corrupt scenarios, robust Dorfman consistently outperformed competitors, exhibiting the lowest RMSE and the highest clinical relevance ratio for selecting genes in NSCLC cell lines whose expression most strongly predicts trametinib IC50. This method, albeit promising, we do acknowledge modifications required for extension to ultra-high dimensions (e.g., $p = 100{,}000$) in variable selection, including improvements in the first step such as block OGK, block graphical LASSO, and LSH in group merging, as well as Bayesian-optimized hyperparameter selection, with the use of HPC clusters.
\printbibliography
\newpage
\appendix
\setcounter{algorithm}{2}

\section{A: Supplementary Figures, performance metrics and competitor methods}

\begin{figure}
    \centering
    \includegraphics[width=\linewidth]{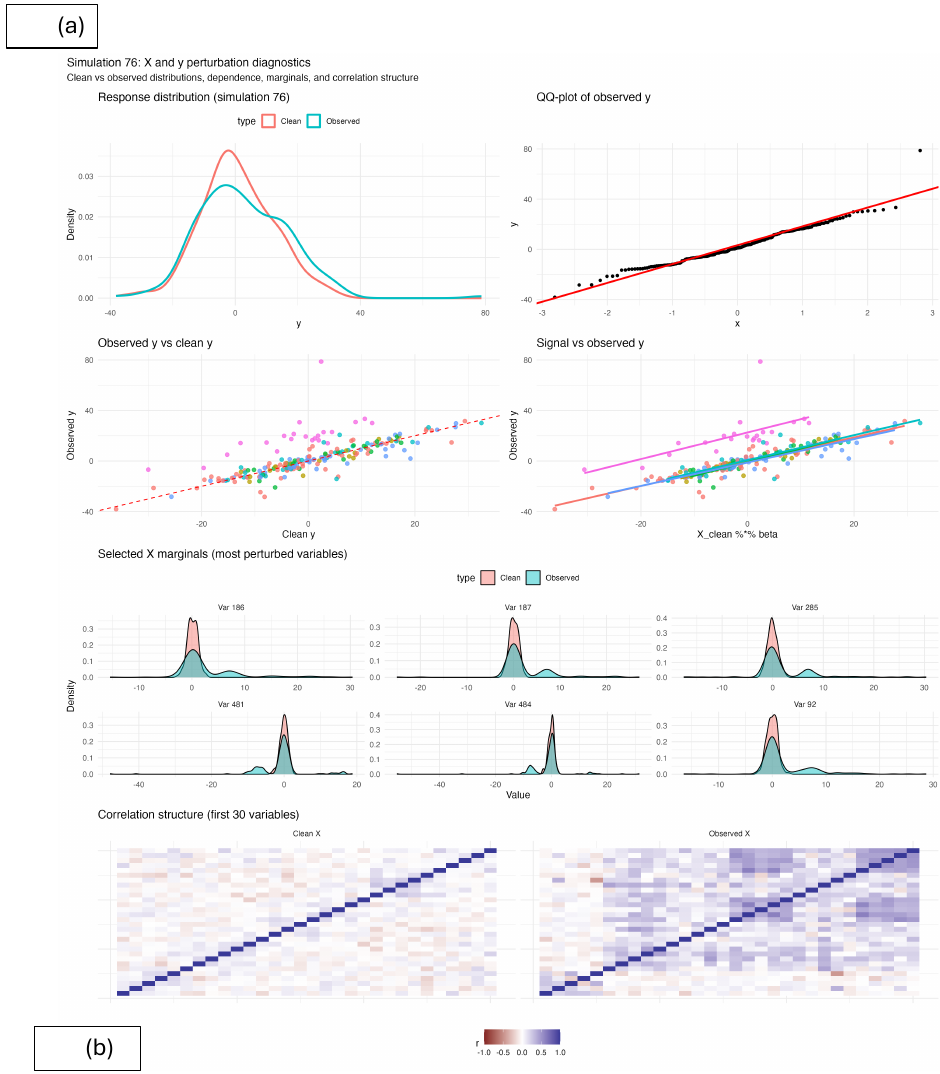}
    \label{fig:a1a}
\end{figure}

\begin{figure}
    \centering
    \includegraphics[width=\linewidth]{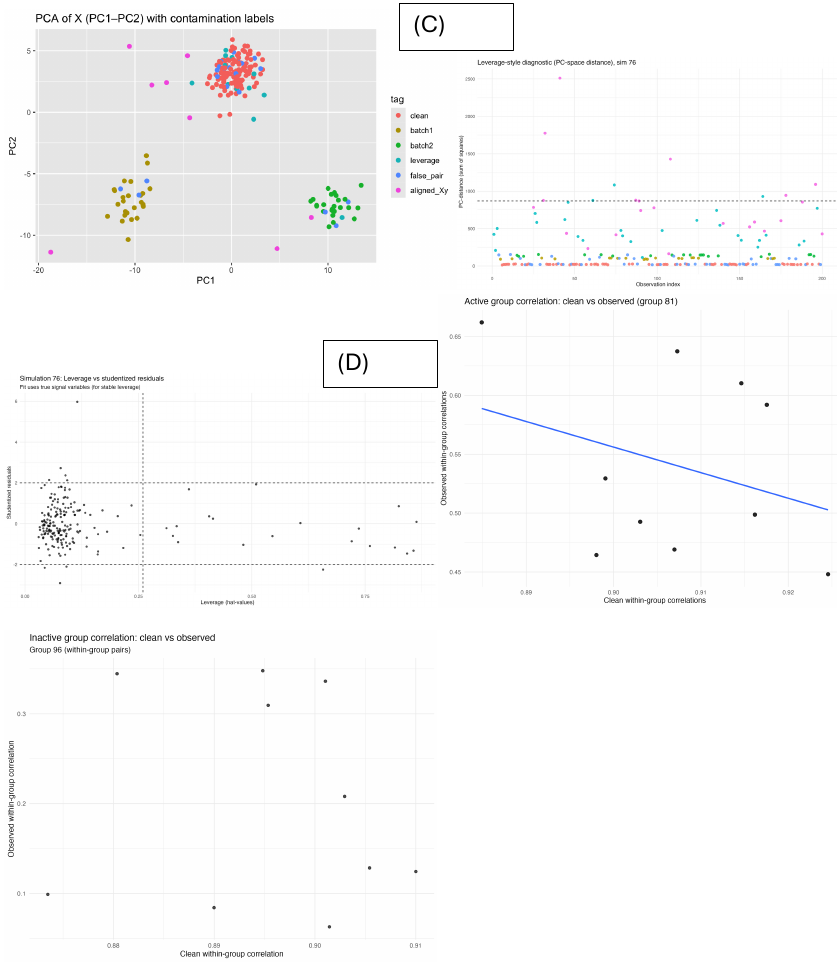}
    \caption{Diagnostic plots of corrupted data. (A) QQ Plots, Skewness and Kurtosis plots, and correlation plots. The corrupted data exhibited outliers, deviations from normality, and homoscedastic errors. The correlations within group for the first 30 variables are shown; in the clean simulation, all correlations are 0.1 (left). However, in the corrupted case, all correlations are in the 0.5-1 range. (B) PCA plot of X in the corrupted simulation, batch effects are clearly visible. (C) High residual and leverage points are clearly shown (D). Both group 81 and 96 have within-group correlation of 0.9; the corrupted correlations are in the range of 0.45-0.65 and 0.1-0.3, respectively. 
}
    \label{fig:a1b}
\end{figure}

\begin{figure}
    \centering
    \includegraphics[width=\linewidth]{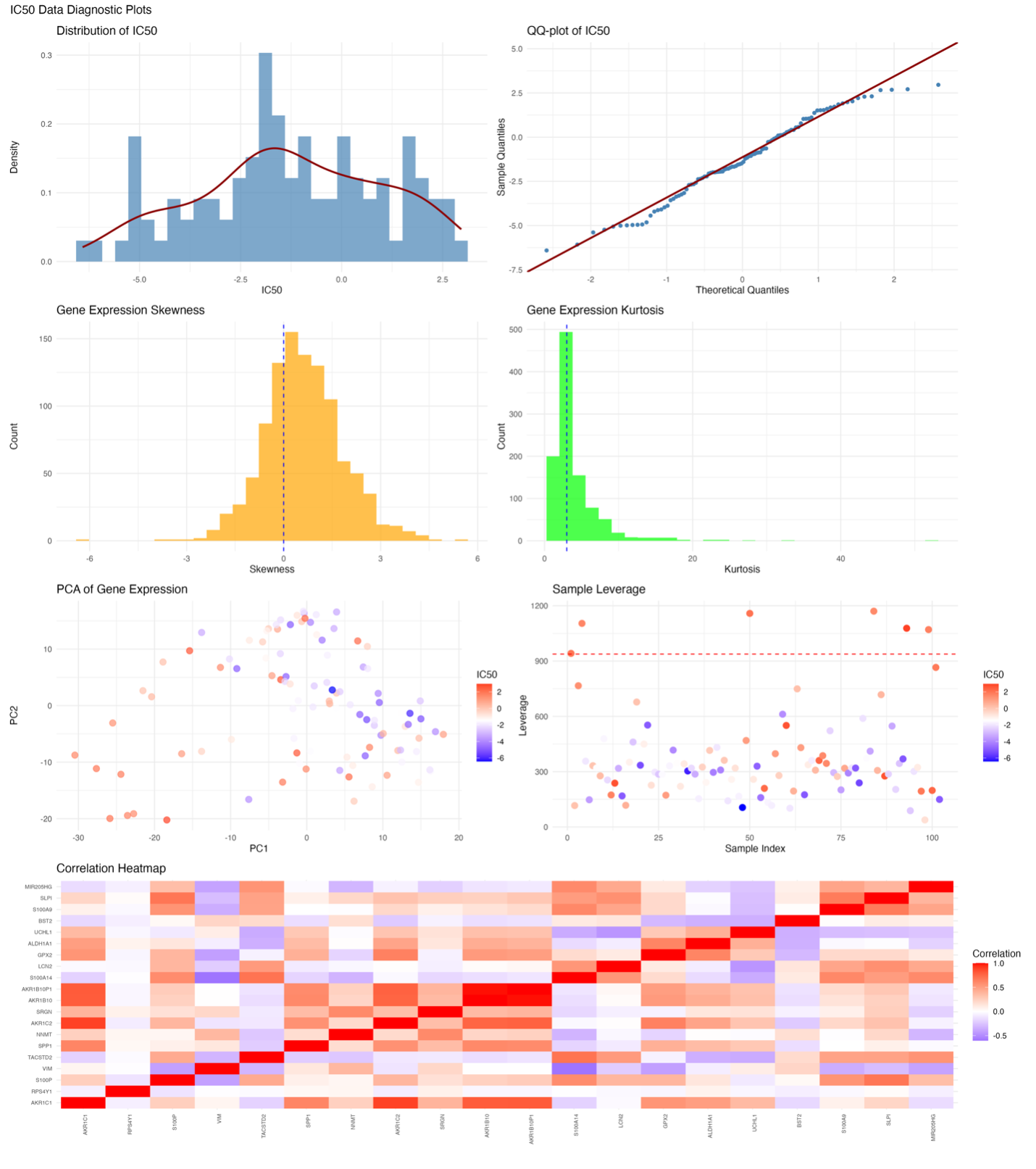}
    \caption{Diagnostic plots of trametinib IC50 data. IC50 distribution and QQ plots clearly show outliers and non-normality. Gene expression data exhibits kurtosis, skewness, and high-leverage points. 
}
    \label{fig:a2}
\end{figure}

\subsection{A.1 Evaluation protocol.}
For each simulation replicate, the $n$ observations were randomly split into a training set $\mathcal{D}_{\mathrm{tr}}$ (60\%) and a testing set $\mathcal{D}_{\mathrm{te}}$ (40\%). All model tuning and variable screening were performed \emph{only} on $\mathcal{D}_{\mathrm{tr}}$. In particular, hyperparameters (e.g., cut height $h$, screening thresholds $\alpha_1,\alpha_2$, and sparsification threshold $\tau$ when applicable) were selected by $K$-fold cross-validation within the training set (with $K=5$), minimizing the average validation RMSE. After tuning, the method was refit on the full training set using the selected hyperparameters to produce a final selected variable set $\widehat{\mathcal{S}}$. Predictive performance was then evaluated on the independent testing set $\mathcal{D}_{\mathrm{te}}$. 

\subsection{A.2 Selection accuracy metrics.}
Let $\mathcal{S}^{\star}=\{j:\beta_j\neq 0\}$ denote the true signal set and $\widehat{\mathcal{S}}$ the selected set. Define
\[
\mathrm{TP}=|\widehat{\mathcal{S}}\cap \mathcal{S}^{\star}|,\qquad
\mathrm{FP}=|\widehat{\mathcal{S}}\setminus \mathcal{S}^{\star}|,\qquad
\mathrm{FN}=|\mathcal{S}^{\star}\setminus \widehat{\mathcal{S}}|.
\]
The true positive rate (TPR, i.e., recall) and false discovery rate (FDR) are
\[
\mathrm{TPR}=\frac{\mathrm{TP}}{\mathrm{TP}+\mathrm{FN}},\qquad
\mathrm{FDR}=\frac{\mathrm{FP}}{\mathrm{TP}+\mathrm{FP}}.
\]
Precision (positive predictive value, PPV) is
\[
\mathrm{PPV}=\frac{\mathrm{TP}}{\mathrm{TP}+\mathrm{FP}}.
\]
The F1 score is the harmonic mean of precision and recall:
\[
\mathrm{F1}
=\frac{2\,\mathrm{PPV}\cdot \mathrm{TPR}}{\mathrm{PPV}+\mathrm{TPR}}
=\frac{2\,\mathrm{TP}}{2\,\mathrm{TP}+\mathrm{FP}+\mathrm{FN}}.
\]
We also report the number of selected variables $|\widehat{\mathcal{S}}|$.

\subsection{A.3 Prediction error (RMSE).}
Let $\hat{f}_{\mathrm{tr}}(\cdot)$ denote the final prediction rule trained on $\mathcal{D}_{\mathrm{tr}}$ using the selected variables $\widehat{\mathcal{S}}$ (in our implementation, ordinary least squares fit on $\mathcal{D}_{\mathrm{tr}}$ restricted to $\widehat{\mathcal{S}}$). The test RMSE is
\[
\mathrm{RMSE}
=\left(\frac{1}{|\mathcal{D}_{\mathrm{te}}|}
\sum_{i\in \mathcal{D}_{\mathrm{te}}}
\big(y_i-\hat{f}_{\mathrm{tr}}(\mathbf{x}_i)\big)^2\right)^{1/2}.
\]
When $\widehat{\mathcal{S}}=\varnothing$, we use the intercept-only predictor
$\hat{f}_{\mathrm{tr}}(\mathbf{x})=\bar{y}_{\mathrm{tr}}$.

\subsection{A.4 Adjusted \texorpdfstring{$R^2$}{R2} and AIC (test-set version).}
Let $n_{\mathrm{te}}=|\mathcal{D}_{\mathrm{te}}|$ and $d=|\widehat{\mathcal{S}}|$. Define
\[
\mathrm{RSS}=\sum_{i\in \mathcal{D}_{\mathrm{te}}}\big(y_i-\hat{f}_{\mathrm{tr}}(\mathbf{x}_i)\big)^2,\qquad
\mathrm{TSS}=\sum_{i\in \mathcal{D}_{\mathrm{te}}}\big(y_i-\bar{y}_{\mathrm{te}}\big)^2,
\]
and
\[
R^2 = 1-\frac{\mathrm{RSS}}{\mathrm{TSS}},\qquad
R^2_{\mathrm{adj}} = 1-\frac{\mathrm{RSS}/(n_{\mathrm{te}}-d-1)}{\mathrm{TSS}/(n_{\mathrm{te}}-1)}.
\]
We report an AIC-style criterion computed on the test residuals:
\[
\mathrm{AIC}=n_{\mathrm{te}}\log\!\left(\frac{\mathrm{RSS}}{n_{\mathrm{te}}}\right)+2d.
\]

\section{B: Bayesian Hyperparameter Optimization}
\addcontentsline{toc}{section}{Appendix A: Bayesian Hyperparameter Optimization}

\subsection{B.1 Bayesian Optimization Framework}
\label{ss:bayes-opt}

Instead of brute-force grid search, we employ Bayesian Optimization (BO) to find the optimal hyperparameters $(\alpha_1, \alpha_2, \rho, h)$. The complete algorithm is as follows:

\subsubsection{B.1.1 Gaussian Process Prior}
We place a Gaussian Process prior on the objective function:
\[
f(\cdot) \sim \mathcal{GP}(\mu(\cdot), k(\cdot, \cdot))
\]
where $\mu(\cdot)$ is the mean function and $k(\boldsymbol{\theta}, \boldsymbol{\theta}')$ is the Mat\'ern kernel.

\subsubsection{B.1.2 Objective Function}
The objective function to minimize is the 5-fold cross-validated RMSE:
\[
f(\boldsymbol{\theta}) = \text{RMSE}_{\text{CV}}(\alpha_1, \alpha_2, \rho, h), \quad \boldsymbol{\theta} = (\alpha_1, \alpha_2, \rho, h) \in \Theta
\]
with $\Theta = [0,1] \times [0,1] \times [0.001, 0.3] \times [0.1, 0.9]$.

\subsubsection{B.1.3 Posterior Update}
After $t$ evaluations, the posterior distribution is updated via Bayes' rule:
\[
p(f|\mathcal{D}_t) \propto p(\mathcal{D}_t|f)p(f)
\]
where $\mathcal{D}_t = \{(\boldsymbol{\theta}_i, f(\boldsymbol{\theta}_i))\}_{i=1}^t$.

\subsubsection{B.1.4 Acquisition Function}
The next evaluation point is selected by maximizing the Expected Improvement (EI):
\[
\boldsymbol{\theta}_{t+1} = \arg\max_{\boldsymbol{\theta} \in \Theta} \text{EI}_t(\boldsymbol{\theta})
\]
where
\[
\text{EI}_t(\boldsymbol{\theta}) = \mathbb{E}[\max(0, f_{\min} - f(\boldsymbol{\theta}))|\mathcal{D}_t]
\]
and $f_{\min} = \min\{f(\boldsymbol{\theta}_1), \dots, f(\boldsymbol{\theta}_t)\}$.

\subsubsection{B.1.5 Algorithm}

Algorithm~3 depicts a typical Bayesian Hyperparameter Optimization design.

\begin{algorithm}[htbp]
\small
\caption{Bayesian Hyperparameter Optimization}
\begin{algorithmic}[1]
\Require Data $(X, y)$, initial points $n_0$, total evaluations $T$
\Ensure Optimal hyperparameters $\boldsymbol{\theta}^*$
\State Initialize $\mathcal{D}_{n_0}$ with $n_0$ random evaluations
\For{$t = n_0$ to $T-1$}
    \State Update GP posterior $p(f|\mathcal{D}_t)$
    \State Compute $\boldsymbol{\theta}_{t+1} = \arg\max_{\boldsymbol{\theta}} \text{EI}_t(\boldsymbol{\theta})$
    \State Evaluate $y_{t+1} = f(\boldsymbol{\theta}_{t+1})$
    \State Update $\mathcal{D}_{t+1} = \mathcal{D}_t \cup \{(\boldsymbol{\theta}_{t+1}, y_{t+1})\}$
\EndFor
\State \Return $\boldsymbol{\theta}^* = \arg\min_{\boldsymbol{\theta} \in \{\boldsymbol{\theta}_1, \dots, \boldsymbol{\theta}_T\}} f(\boldsymbol{\theta})$
\label{alg:bo}
\end{algorithmic}
\end{algorithm}

\subsection{B.2 Computational Advantage}
Bayesian Optimization requires only 10--50$\times$ fewer evaluations compared to exhaustive grid search while achieving comparable or better performance. This makes it particularly suitable for computationally expensive pipelines, such as the Dorfman method, in which each evaluation involves 5-fold cross-validation.

\subsection{B.3 Scalable Hyperparameter Optimization for Ultra-High Dimensions} 

For $p=100,000$, $n=200$, full Bayesian optimization over $(\alpha_1, \alpha_2, \rho, h)$ is computationally prohibitive due to the cost of each Dorfman pipeline evaluation. We therefore employ a \textbf{multi-fidelity Bayesian optimization} approach:

\begin{enumerate}
    \item \textbf{Low-fidelity screening stage:} Optimize $(\alpha_1, \alpha_2)$ using only the screened subset ($d=5,000$ variables):
   \[
   f_{\text{low}}(\alpha_1, \alpha_2) = \text{RMSE}_{\text{CV}}(\alpha_1, \alpha_2 \mid \mathbf{X}^{(S)})
   \]
   where $\mathbf{X}^{(S)}$ is the top $5,000$ screened variables. We use standard Bayesian optimization (Section~B.1) for this 2D problem.

   \item \textbf{Multi-resolution $\rho$ optimization:} For graphical lasso penalty $\rho$:
   a. \textbf{Coarse grid:} Test $\rho \in \{0.01, 0.05, 0.1\}$ on small subsamples ($n'=100$, $p'=1,000$)
   b. \textbf{Fine-tuning:} Use Bayesian optimization on the promising region identified in (a)

   \item \textbf{Hierarchical $h$ optimization:} For dendrogram cut height $h$:
   \[
   h_{\text{opt}} = \text{argmin}_h \left\{ \text{RMSE}(h) + \lambda \cdot \text{Complexity}(h) \right\}
   \]
   where $\text{Complexity}(h)$ penalizes too many groups (to avoid overfitting).

   \item \textbf{Sequential optimization schedule:}
   \begin{algorithmic}[1]
   \State Fix $\rho = \rho_0$, optimize $(\alpha_1, \alpha_2)$ via BO (10 evaluations)
   \State Fix $(\alpha_1, \alpha_2)$, optimize $\rho$ via multi-resolution BO (8 evaluations)
   \State Fix $(\alpha_1, \alpha_2, \rho)$, optimize $h$ via grid search (5 values)
   \State Joint fine-tuning of all parameters (5 evaluations)
   \end{algorithmic}

   \item \textbf{Parallel evaluation via subsetting:} Each BO evaluation uses: (a) $80\%$ data subsample (n=160); (b), $10\%$ variable subsample (p=10,000 after screening); and (c) Early stopping if RMSE > current best + margin. \textbf{Total evaluations:} 28 vs. 600+ for full grid search.

   \item \textbf{Adaptive resource allocation:} For iteration $t$, allocate computational budget:
   \[
   \text{Budget}_t \propto \exp(-\eta \cdot \text{EI}_t(\theta_t))
   \]
   where $\eta$ controls the exploration-exploitation trade-off.
\end{enumerate}

This approach reduces the computational cost from $O(10^3)$ Dorfman evaluations to $O(10^1)$ while maintaining optimization quality.

\subsection{B.4 Clustering Algorithms for Ultra-High Dimensions} 

\subsubsection{B.4.1 Normal Case for Ultra-High Dimensions ($p=100,000$, $n=200$)}

\textbf{Setting:} $n=200$, $p=100,000$ ($p/n=500$, extreme ultra-high dimension). In this regime, full covariance estimation is computationally infeasible and statistically inconsistent. We employ a scalable three-stage screening, blocking, and merging approach.

\begin{enumerate}
    \item \textbf{Marginal screening:} Perform marginal correlation screening to reduce dimension:
   \[
   \hat{r}_j = \text{corr}(X_j, y), \quad j=1,\dots,p
   \]
   Retain the top $d = \min(5000, 2n) = 400$ covariates with largest $|\hat{r}_j|$, denoted $\mathbf{X}^{(S)} \in \mathbb{R}^{n \times d}$.

   \item \textbf{Blockwise adaptive graphical lasso:} Partition the screened variables into $K = \lceil d/100 \rceil = 40$ blocks of size approximately 100:
   \[
   \mathbf{X}^{(S)} = [\mathbf{X}_1, \dots, \mathbf{X}_K]
   \]
   For each block $k=1,\dots,K$, in parallel:
   
   a. Compute within-block covariance $\hat{\Sigma}^{(k)}$.
   
   b. For $\rho \in \mathcal{R} = \{0.1, 0.2, 0.3, 0.4\}$, fit blockwise adaptive graphical lasso:
   \[
   \hat{\Theta}^{(k)}(\rho) = \arg\min_{\Theta \succ 0} \left\{ \operatorname{tr}(\hat{\Sigma}^{(k)}\Theta) - \log\det(\Theta) + \rho \sum_{i \neq j} w_{ij}^{(k)} |\Theta_{ij}| \right\},
   \]
   where $w_{ij}^{(k)} = \exp(-\gamma |\hat{R}_{ij}^{(k)}|)$ with $\gamma=2$.
   
   c. Compute sparse correlation matrix $\hat{R}^{(k)}(\rho)$ from $\hat{\Theta}^{(k)}(\rho)$.
   
   d. Perform hierarchical clustering on $D^{(k)}(\rho) = 1 - |\hat{R}^{(k)}(\rho)|$ to obtain within-block groups $\mathcal{G}^{(k)} = \{G_1^{(k)}, \dots, G_{m_k}^{(k)}\}$.

   \item \textbf{Between-block grouping via Locality-Sensitive Hashing (LSH):} To efficiently merge groups across blocks without exhaustive pairwise comparisons:
   
   a. \textbf{Compute group signatures:} For each group $G$, compute a signature vector:
   \[
   \text{signature}(G) = \begin{bmatrix}
   \text{mean}(|\hat{r}_j| \text{ for } j \in G) \\
   \text{max}(|\hat{r}_j| \text{ for } j \in G) \\
   \text{std}(|\hat{r}_j| \text{ for } j \in G) \\
   \text{size}(G) / d
   \end{bmatrix}
   \]
   
   b. \textbf{Hash groups into buckets:} Apply $L=3$ independent hash functions $h_\ell: \mathbb{R}^4 \to \{1,\dots,B\}$ with $B=100$ buckets:
   \[
   h_\ell(\mathbf{s}) = \left\lfloor \frac{\mathbf{a}_\ell^\top \mathbf{s} + b_\ell}{w} \right\rfloor \mod B
   \]
   where $\mathbf{a}_\ell \sim \mathcal{N}(0,I_4)$, $b_\ell \sim \text{Uniform}(0,w)$, and $w$ is the bucket width. Groups are placed in the same bucket if \textbf{any} $h_\ell$ matches.
   
   c. \textbf{Merge within buckets:} For each bucket containing groups $\{G_1, \dots, G_t\}$:
   \begin{itemize}
   \item Sort groups by maximum $|\hat{r}_j|$ within group
   \item For each pair $(G_a, G_b)$ from different blocks:
     \[
     \tilde{R}_{ab} = \frac{\hat{r}_{a}^{\max} \cdot \hat{r}_{b}^{\max} \cdot \hat{\sigma}_y^2 + \epsilon}{1+\lambda}
     \]
     where $\hat{r}_{a}^{\max} = \max_{j \in G_a} |\hat{r}_j|$, $\hat{\sigma}_y^2$ is response variance, $\lambda=0.1$, $\epsilon \sim \mathcal{N}(0,0.01)$.
   \item Merge if $|\tilde{R}_{ab}| > \tau = 0.7$
   \end{itemize}
   
   This reduces comparisons from $O(m^2)$ to $O(m)$ where $m \approx 400$ is total number of groups.

   \item \textbf{Parallel parameter selection:} Select $\rho^\star$ via distributed cross-validation across blocks, maximizing average Adjusted Rand Index (ARI). The final groups $\mathcal{G}^{\text{final}}$ are obtained after LSH-based merging.
\end{enumerate}

The resulting groups are sparse, computationally tractable, and statistically regularized for $p=100,000$, $n=200$. The LSH approach ensures efficient cross-block merging without exhaustive pairwise comparisons.

\subsubsection{B.4.2 Robust Case for Ultra-High Dimensions with Contamination ($p=100,000$, $n=200$)}

\textbf{Setting:} $n=200$, $p=100,000$ with potential outliers. We combine robust screening with distributed estimation and LSH-based merging.

\begin{enumerate}
    \item \textbf{Robust marginal screening:} Compute robust marginal associations:
   \[
   \hat{\tau}_j = \text{Kendall's} \tau(X_j, y), \quad j=1,\dots,p
   \]
   Retain the top $d = \min(3000, 1.5n) = 300$ covariates with largest $|\hat{\tau}_j|$, denoted $\mathbf{X}^{(S)}_{\text{rob}}$.

   \item \textbf{Distributed robust graphical lasso:} Partition into $K = \lceil d/50 \rceil = 60$ blocks of size ~50:
   \[
   \mathbf{X}^{(S)}_{\text{rob}} = [\mathbf{X}_1^{\text{rob}}, \dots, \mathbf{X}_K^{\text{rob}}]
   \]
   For each block $k$, in parallel:
   
   a. Compute robust covariance via median-of-medians OGK:
   \[
   \hat{\Sigma}^{(k)}_{\text{rob}} = \text{medOGK}(\mathbf{X}_k^{\text{rob}})
   \]
   
   b. For $\rho \in \mathcal{R} = \{0.2, 0.3, 0.4, 0.5\}$, fit graphical lasso:
   \[
   \hat{\Theta}^{(k)}_{\text{rob}}(\rho) = \arg\min_{\Theta \succ 0} \left\{ \operatorname{tr}(\hat{\Sigma}^{(k)}_{\text{rob}}\Theta) - \log\det(\Theta) + \rho \|\Theta\|_1 \right\}
   \]
   
   c. Compute robust sparse correlation $\hat{R}^{(k)}_{\text{rob}}(\rho)$.
   
   d. Perform dynamic cut HC on $D^{(k)}_{\text{rob}}(\rho) = 1 - |\hat{R}^{(k)}_{\text{rob}}(\rho)|$.

   \item \textbf{Robust LSH for cross-block merging:}
   
   a. \textbf{Robust signatures:} For each robust group $G^{\text{rob}}$:
   \[
   \text{signature}_{\text{rob}}(G^{\text{rob}}) = \begin{bmatrix}
   \text{median}(|\hat{\tau}_j| \text{ for } j \in G^{\text{rob}}) \\
   \text{mad}(|\hat{\tau}_j| \text{ for } j \in G^{\text{rob}}) \\
   \text{size}(G^{\text{rob}}) / d \\
   \text{robust skewness}
   \end{bmatrix}
   \]
   where mad is median absolute deviation.
   
   b. \textbf{Robust hashing:} Use cosine similarity LSH for robustness:
   \[
   h^{\text{rob}}_\ell(\mathbf{s}) = \text{sign}(\mathbf{a}_\ell^\top \mathbf{s})
   \]
   where $\mathbf{a}_\ell \sim \mathcal{N}(0,I_4)$. Groups with matching hash signatures across multiple $\ell$ are placed in same bucket.
   
   c. \textbf{Robust merging:} Within each bucket, merge groups $G_a^{\text{rob}}$ and $G_b^{\text{rob}}$ if:
   \[
   \min\left(\text{median}_{j \in G_a^{\text{rob}}} |\hat{\tau}_j|, \text{median}_{j \in G_b^{\text{rob}}} |\hat{\tau}_j|\right) > 0.6
   \]
   and they originate from different blocks.

   \item \textbf{Consensus clustering:} Repeat steps 2-3 with different random partitions, then apply consensus clustering for stability.

   \item \textbf{Dynamic tree cutting:} Apply multi-level dynamic tree cut with deepSplit=2 to automatically determine final group hierarchy.
\end{enumerate}

\textbf{Computational complexity:} The LSH-based approach reduces cross-block comparisons from $O(K^2 \cdot g^2)$ to $O(K \cdot g \cdot L)$ where $K=60$ blocks, $g \approx 5$ groups per block, and $L=3$ hash functions, making it feasible for $p=100,000$.

\section{C: Breakdown of Elastic Net and Adaptive Elastic Net in Ultra-High Dimensions}

\subsection{C.1 Theoretical Framework}

Consider the linear model:
\[
y = X\beta^* + \varepsilon, \quad \varepsilon \sim \mathcal{N}(0, \sigma^2 I_n)
\]
where $X \in \mathbb{R}^{n \times p}$, $p \gg n$, and $\beta^*$ is $s$-sparse ($\|\beta^*\|_0 = s$).

\subsection{C.2 Standard Elastic Net (EN) Breakdown}

The EN estimator solves:
\[
\hat{\beta}_{\text{EN}} = \arg\min_{\beta} \left\{ \frac{1}{2n}\|y - X\beta\|_2^2 + \lambda_1 \|\beta\|_1 + \lambda_2 \|\beta\|_2^2 \right\}
\]

\textbf{Breakdown occurs when $p/n \to \infty$ due to:}

\begin{enumerate}
    \item \textbf{Irrepresentable Condition Failure:}
\[
\max_{j \notin S} \|X_j^\top X_S (X_S^\top X_S)^{-1}\|_1 \geq 1
\]
where $S = \text{supp}(\beta^*)$. For $p \gg n$, with high probability:
\[
\mathbb{P}\left(\max_{j \notin S} \|X_j^\top X_S (X_S^\top X_S)^{-1}\|_1 > 1 + \delta\right) \to 1
\]
for some $\delta > 0$, violating the condition required for consistent variable selection.

    \item \textbf{Noise Accumulation:}
\[
\frac{\|X^\top \varepsilon\|_\infty}{n} = O_P\left(\sqrt{\frac{\log p}{n}}\right) \to \infty \quad \text{when } \frac{\log p}{n} \to \infty
\]
This leads to false positives because noise covariates appear spuriously correlated with the residuals.

    \item \textbf{Collinearity Amplification:}
\[
\kappa(X^\top X) = O\left(\frac{p}{n}\right) \to \infty
\]
where $\kappa(\cdot)$ is the condition number, leading to unstable coefficient estimates.
\end{enumerate}

\subsection{C.3 Adaptive Elastic Net (AdaEN) Breakdown}

The AdaEN estimator uses initial weights $\hat{w}_j = |\hat{\beta}^{(0)}_j|^{-\gamma}$:
\[
\hat{\beta}_{\text{AdaEN}} = \arg\min_{\beta} \left\{ \frac{1}{2n}\|y - X\beta\|_2^2 + \lambda \sum_{j=1}^p \hat{w}_j |\beta_j| \right\}
\]

\textbf{Breakdown is more severe due to:}

\begin{enumerate}
    \item \textbf{Initial Estimator Inconsistency:} When $p \gg n$, any initial estimator $\hat{\beta}^{(0)}$ (e.g., marginal correlation, ridge) satisfies:
    \[
    \|\hat{\beta}^{(0)} - \beta^*\|_2 = O_P\left(\sqrt{\frac{s \log p}{n}}\right) \to \infty
    \]
    making weights $\hat{w}_j$ unreliable.

    \item \textbf{Weight Inversion Problem:}
    For truly zero coefficients ($\beta_j^* = 0$), with high probability:
    \[
    |\hat{\beta}^{(0)}_j| = O_P\left(\sqrt{\frac{\log p}{n}}\right) \quad \Rightarrow \quad \hat{w}_j = O_P\left(\left(\frac{n}{\log p}\right)^{\gamma/2}\right) \to \infty
    \]
    Thus, truly irrelevant variables receive \textbf{vanishing penalties}, increasing false positives.

    \item\textbf{Signal Attenuation:}
    For truly non-zero coefficients ($\beta_j^* \neq 0$):
    \[
    |\hat{\beta}^{(0)}_j| = |\beta_j^*| + O_P\left(\sqrt{\frac{\log p}{n}}\right) \quad \Rightarrow \quad \hat{w}_j = |\beta_j^*|^{-\gamma} + O_P\left(\frac{1}{\sqrt{n \log p}}\right)
    \]
    Signal variables receive \textbf{excessive penalties} relative to noise variables, causing false negatives.
\end{enumerate}

\subsection{C.4 Quantitative Breakdown Conditions}

\textbf{Theorem C.1 (EN Breakdown):} For $p = O(\exp(n^\alpha))$ with $\alpha > 0$, EN fails to achieve consistent selection if:
\[
\frac{\log p}{n} \to \infty \quad \text{or} \quad \frac{s \log p}{n} \to \infty
\]

\textbf{Theorem C.2 (AdaEN Accelerated Breakdown):} AdaEN breaks down under weaker conditions:
\[
\frac{\log p}{\sqrt{n}} \to \infty \quad \text{or} \quad \frac{s \sqrt{\log p}}{n} \to \infty
\]
The breakdown is more severe due to miscalibration of weight.

\subsection{C.5 Adaptive PENSE Breakdown in Ultra-High Dimensions}

Adaptive PENSE (Penalized Elastic Net S-Estimator) combines robust S-estimation with adaptive elastic net regularization. The estimator is defined as:

\[
\hat{\beta}_{\text{AdaPENSE}} = \arg\min_{\beta} \left\{ \hat{\sigma}^2(\beta) + \lambda \sum_{j=1}^p \hat{w}_j |\beta_j| \right\}
\]

where $\hat{\sigma}(\beta)$ is an S-estimator of scale satisfying:
\[
\frac{1}{n} \sum_{i=1}^n \rho\left(\frac{y_i - x_i^\top \beta}{\hat{\sigma}(\beta)}\right) = b
\]
with $\rho$ a symmetric, bounded loss function (e.g., Tukey's biweight), and weights $\hat{w}_j = |\hat{\beta}^{(0)}_j|^{-\gamma}$ computed from an initial robust estimator.

\textbf{Breakdown in ultra-high dimensions ($p \gg n$) occurs through three mechanisms:}

\begin{enumerate}
    \item \textbf{Robust Initial Estimator Failure:}
   The initial robust estimator $\hat{\beta}^{(0)}$ (typically MM-estimator or S-estimator on screened variables) becomes inconsistent when $p/n \to \infty$:
   \[
   \|\hat{\beta}^{(0)} - \beta^*\|_2 = O_P\left(\sqrt{\frac{s \log p}{n}}\right) \to \infty
   \]
   This inconsistency propagates through the adaptive weights $\hat{w}_j$, resulting in the same weight-inversion problem as in classical AdaEN.

    \item \textbf{S-Estimation Breakdown Point Reduction:}
   The finite-sample breakdown point $\epsilon^*_n$ of S-estimators degrades with dimension:
   \[
   \epsilon^*_n = \min\left(\frac{\lfloor (n-p)/2 \rfloor + 1}{n}, \frac{\lfloor p/2 \rfloor + 1}{n}\right)
   \]
   For $p > n$, this gives $\epsilon^*_n = \frac{\lfloor (n-p)/2 \rfloor + 1}{n} < 0$, indicating \textbf{no robustness guarantee} in ultra-high dimensions.

    \item \textbf{Scale Estimation Instability:}
   The S-estimator of scale $\hat{\sigma}(\beta)$ loses consistency:
   \[
   \left|\frac{\hat{\sigma}(\beta^*)}{\sigma} - 1\right| = O_P\left(\sqrt{\frac{p}{n}}\right) \to \infty
   \]
   This inflates the objective function and destabilizes optimization.
\end{enumerate}

\textbf{Key Limitation:} Adaptive PENSE assumes $p < n$ for its theoretical robustness guarantees. When $p \gg n$, its breakdown point reduces to:
\[
\epsilon^*_{\text{effective}} = \frac{1}{2} \cdot \frac{n}{p} \to 0
\]
For $p=100,000$, $n=200$, this gives $\epsilon^*_{\text{effective}} \approx 0.001$, offering virtually no robustness protection.

\noindent\textbf{Corollary C.4 (PENSE Ultra-High Dimensional Breakdown):} For $p = O(\exp(n^\alpha))$ with $\alpha > 0$, adaptive PENSE loses its oracle properties and robustness guarantees. The breakdown occurs earlier than for classical methods due to the combined challenges of robust estimation and weight adaptation in high dimensions.

This explains why even state-of-the-art robust regularized methods, such as adaptive PENSE, require modifications (screening, blocking) for ultra-high-dimensional settings, as implemented in our robust Dorfman pipeline.

\subsection{C.6 Dorfman's Advantage in Ultra-High Dimensions}

The Dorfman method mitigates these issues through:

\begin{enumerate}
    \item \textbf{Dimensionality Reduction:}
    \[
    p \xrightarrow{\text{screening}} d = O(n) \xrightarrow{\text{grouping}} K = O(\sqrt{n})
    \]
    reducing the effective dimension before penalization.
    
    \item \textbf{Hierarchical Testing:}
    Two-stage testing with $\alpha_1, \alpha_2$ controls error propagation:
    \[
    \mathbb{P}(\text{false selection}) \leq \alpha_1 + (1-\alpha_1)\alpha_2
    \]

    \item \textbf{Group Sparsity:}
    Exploits $g$-group sparsity rather than $s$-element sparsity:
    \[
    \|\beta^*\|_{2,0} \ll \|\beta^*\|_0
    \]
    where $\|\beta\|_{2,0} = \sum_{g=1}^G \mathbb{I}(\|\beta_g\|_2 > 0)$.
\end{enumerate}

\noindent\textbf{Corollary C.3:} Under group sparsity with $G = O(n/\log n)$, Dorfman achieves:
\[
\mathbb{P}(\hat{S} = S) \to 1 \quad \text{when} \quad \frac{G \log G}{n} \to 0
\]
even as $p/n \to \infty$, provided screening retains relevant groups.

\subsection{C.7 Implications for Method Choice}

The breakdown analysis suggests:
\begin{itemize}
\item \textbf{Standard EN:} Avoid when $p/n > 50$ without preprocessing
\item \textbf{AdaEN:} Particularly dangerous when $p/n > 10$ due to weight instability
\item \textbf{Dorfman:} Recommended for $p/n > 100$ when group structure exists
\item \textbf{Screening necessity:} Essential for $p > 1000n$ regardless of method
\end{itemize}

This justifies our methodological choices for ultra-high-dimensional settings.

\end{document}